\documentclass[pdflatex, sn-basic]{sn-jnl}

\usepackage{graphicx}%
\usepackage{multirow}%
\usepackage{amsmath,amssymb,amsfonts}%
\usepackage{amsthm}%
\usepackage{mathrsfs}%
\usepackage[title]{appendix}%
\usepackage{xcolor}%
\usepackage{textcomp}%
\usepackage{manyfoot}%
\usepackage{booktabs}%
\usepackage{algorithm}%
\usepackage{algorithmicx}%
\usepackage{algpseudocode}%
\usepackage{listings}%

\usepackage{amssymb}
\usepackage[figuresright]{rotating}
\usepackage{hyperref}
\usepackage{booktabs}
\usepackage{amsmath}
\usepackage{color, colortbl}
\definecolor{light-gray}{gray}{0.95}

\usepackage[inline]{enumitem}
\usepackage{algorithm}
\usepackage{algpseudocode}  
\usepackage{adjustbox}
\usepackage{caption}
\usepackage{rotating}

\begin{document}


\title{Personalized Federated Learning for improving radar based precipitation nowcasting on heterogeneous areas}


\author*[1]{Judith Sáinz-Pardo Díaz}
\email{sainzpardo@ifca.unican.es}

\author[1]{María Castrillo}
\email{castrillo@ifca.unican.es}

\author[2]{Juraj Bartok}
\email{juraj.bartok@microstep-mis.com}

\author[1]{Ignacio Heredia Cachá}
\email{iheredia@ifca.unican.es}

\author[2]{Irina Malkin Ondík}
\email{irina.malkin.ondik@microstep-mis.com}

\author[2]{Ivan Martynovskyi}
\email{ivan.martynovskyi@microstep-mis.com}

\author[3]{Khadijeh Alibabaei}
\email{khadijeh.alibabaei@kit.edu}

\author[3]{Lisana Berberi}
\email{lisana.berberi@kit.edu}

\author[3]{Valentin Kozlov}
\email{valentin.kozlov@kit.edu}

\author[1]{Álvaro López García}
\email{aloga@ifca.unican.es}

\affil[1]{\orgname{Instituto de Física de Cantabria (IFCA), CSIC-UC.}, \orgaddress{\street{Avda. los Castros s/n.}, \city{Santander}, \postcode{39005}, \country{Spain}}}

\affil[2]{\orgname{MicroStep-MIS}, \orgaddress{\street{Čavojského 1}, \city{Bratislava}, \postcode{841 04}, \country{Slovakia}}}

\affil[3]{\orgdiv{Scientific Computing Center (SCC)}, \orgname{Karlsruhe Institute of Technology (KIT)}, \orgaddress{\city{Karlsruhe}, \country{Germany}}}


\abstract{\footnotetext{\textit{Accepted for publication in Earth Science Informatics}}The increasing generation of data in different areas of life, such as the environment, highlights the need to explore new techniques for processing and exploiting data for useful purposes. In this context, artificial intelligence techniques, especially through deep learning models, are key tools to be used on the large amount of data that can be obtained, for example, from weather radars. In many cases, the information collected by these radars is not open, or belongs to different institutions, thus needing to deal with the distributed nature of this data. In this work, the applicability of a personalized federated learning architecture, which has been called \textit{adapFL}, on distributed weather radar images is addressed. To this end, given a single available radar covering 400 km in diameter, the captured images are divided in such a way that they are disjointly distributed into four different federated clients. The results obtained with \textit{adapFL} are analyzed in each zone, as well as in a central area covering part of the surface of each of the previously distributed areas. The ultimate goal of this work is to study the generalization capability of this type of learning technique for its extrapolation to use cases in which a representative number of radars is available, whose data can not be centralized due to technical, legal or administrative concerns. The results of this preliminary study indicate that the performance obtained in each zone with the \textit{adapFL} approach allows improving the results of the federated learning approach, the individual deep learning models and the classical Continuity Tracking Radar Echoes by Correlation approach.}




\maketitle

\section{Introduction and motivation}\label{sec:intro}
Privacy issues and legal restrictions that apply to the field of images and their processing through machine and deep learning (ML and DL) models are clear, for example, in the medical field, where it is essential to preserve the privacy of patients. In this case, it happens that in many cases the images taken in a certain center, such as a hospital, cannot be centralized together with images taken by other institutions. However, this problem regarding data centralization can be due to other reasons and in other areas, such as those related to data storage capacity, connectivity restrictions, or even a lack of computational resources that prevent the training of models on large amounts of data, making decentralization more convenient. In this context, federated learning (FL) architectures allow training ML/DL models in a distributed way, without having to centralize the data in a central server \citep{pmlr-v54-mcmahan17a}. These architectures achieve robustness by aggregating models trained individually by each data owner or client.

This work is focused on images captured by weather radars. Nowadays, there is an increasing demand for weather forecasts with high spatial accuracy, which is boosting private weather forecasting products, e.g. through the installation of private infrastructure such as radars. Their distributed nature, as well as the often proprietary nature of the data, makes it an ideal fit for the FL approach. In addition, high resolution radars produce large data volumes, which are inconvenient to share in order to train a centralized model.

In particular, weather radars are an effective tool for precipitation nowcasting, whose frequency and intensity is expected to be affected by climate change, as well as various other types of extreme weather events \citep{Svoboda2016, Rajczak2017, Hanel2010, Hossein2020}. There are several references that provide clear evidence, such as a fatal flash flood \citep{Svoboda1998} and the deadliest European tornado since 2001 \citep{Korosec2021, Komjati2022}. 

In this work, we develop a novel personalized federated learning (PFL) approach due to the inherent heterogeneous nature of radar-based precipitation data. The aim of PFL is to address the main challenges that faces FL, like its poor convergence on highly heterogeneous data and its lack of generalization ability beyond the global distribution of the data \citep{tan2022towards}. For this study, the images captured by the radar are divided in four quadrants, simulating a use case in which each radar would cover a sub-area of the total region of interest. We study the predictive and generalization capacity of the resulting models, in order to assess the applicability of these methods to real-world use cases, in which data comes from different radars potentially located in different countries. The motivation for addressing this study is two-fold: (1) to enhance collaboration between institutions handling such data without sharing it, (2) to study if training models under a personalized federated learning architecture provide better results than training with individual data from each radar or area of capturing. This is of particular interest when data cannot be centralized due to technical, computational or storage restrictions. In addition, we can mobilize computing wherever data is available, even using isolated GPUs or compute nodes for this task.

The remainder of this work is structured as follows: Section~\ref{sec:sota} presents the state of the art regarding the research areas concerning artificial intelligence models and meteorology, together with the recent state of the art concerning the application of FL in different use cases and PFL. Section~\ref{sec:data} summarizes the data used during this study, as well as the distribution carried out for the purpose of applying the FL and PFL approaches. Section~\ref{sec:methodology} describes the methodology implemented for the development of the models used to carry out the predictions, together with the proposal of a novel PFL architecture (namely adaptive FL, \textit{adapFL}). Section~\ref{sec:results} presents the results, together with their analysis. In Section~\ref{sec:discussion} a discussion concerning different limitations of the study and the data used is presented. Finally, Section~\ref{sec:conclusions} draws the conclusions together with future lines of work.

\section{State of the art}\label{sec:sota}
The state of the art in relation to image processing, meteorology-oriented DL, precipitation nowcasting using DL models, as well as FL and PFL applications is discussed and evaluated in this section.
\subsection{Deep learning techniques for meteorology}

Concerning AI-based models and techniques, specifically the development of DL models, there are several fields that are relevant for the meteorological domain. First, image processing focuses on the analysis of digital images and videos by software programs from statistically based approaches \citep{Tanaka2002, Kim2005} to modern neural networks \citep{GuWang2018}. This field is further subdivided into multiple subdomains, such as object detection, object tracking, object recognition, etc. The aim of object detection is to identify and localize relevant objects from a digital image. Convolutional neural networks, first applied in 1988 to alphabet recognition \citep{Zhang1988} are especially well suited for this task. Specifically, object tracking and recognition aims to identify an object in a sequence of images and can attempt to predict its movement. A recent paper \citep{Kesa2022} describes how to solve both of these tasks at the same time. Siamese neural networks (e.g., DaSiamRPN \citep{ZhuWang2018}, Cascaded SiamRPN \citep{FanLing2018}, SiamMask \citep{WangZhang2019}, SiamRPN++ \citep{LiWu2018}, Deeper and Wider SiamRPN \citep{ZhangPeng2019}), are also useful in object tracking.  



Regarding meteorology-oriented DL, blurry predictions are a tremendous challenge  \citep{Pavlik2022}, requiring a special treatment of loss functions. Generative Adversarial Network (GAN) \citep{agrawal2019machine, Goodfellow2014} has been recently successfully applied to solve this challenge. The approach called DGMR \citep{Ravuri2021} is currently the state-of-the-art in the radar-based storm nowcasting. 

Concerning precipitation nowcasting using DL models, several works analyze the use of different models to better accomplish this task. In \cite{NIPS2017_a6db4ed0} the authors propose the use of the \textit{Trajectory GRU} model instead of the classic Convolutional LSTM (ConvLSTM) which was used for example in \cite{shi2015convolutional}. In addition, in \cite{chen2020deep} the authors also propose a novel deep learning neural network and they start presenting the COTREC method (also compared in this study) and the ConvLSTM. They compare the performance of the different models on composite reflectivity data concerning the critical success index (CSI) over 30 and 60 min. In \cite{KO2022105072} a novel loss function is also used to mitigate the class imbalance problem in the CSI in the case of heavy rainfall. In \cite{9354430} a Deep Transfer Learning approach is proposed for radar nowcasting using a CNN as benchmark and two transfer learning models with few data belonging to the target area (data from 5 days regarding the target area for training and 3 days for testing). Finally, \cite{cuomo2021use} presents different DL models for weather radar nowcasting together with some drawbacks encountered, highlight in Table 2 a review of ML models used for weather nowcasting from the literarure.

\subsection{Federated Learning and Personalized Federated Learning}
With regard to the use of FL in data analysis, the use cases addressed in the state of the art are quite broad. In particular, numerous studies have been conducted in the medical field, due to the obvious privacy restrictions that apply in such environments \citep{10.1145/3412357, rieke2020future, sainzpardo2023fl}. Other applications deal with intrusion detection systems \citep{AGRAWAL2022346}, credit card fraud detection \citep{yang2019ffd}, but also applications that arise when dealing with wireless communications applications \citep{9141214}. Concerning climate sciences and more particularly the study of water quality, the classical FL architecture is applied in \cite{SAINZPARDODIAZ2023120726} to predict the concentration of chlorophyll given different physico-chemical and meteorological features, using two different sites of data gathering, specifically two tributaries of the River Thames. However, to our knowledge, there is not any study in which FL or PFL architectures have been applied to try to improve the predictions from classical models using meteorological radar images.

PFL includes a wide group of different strategies to further tailor an FL model to better fit each client's data, while maintaining privacy and security \citep{SABAH2024122874}. According to the taxonomy proposed in \cite{tan2022towards}, PFL approaches can be classified in two main types: Global Model Personalization and Learning Personalized Models. The approach presented in this work falls into the first group because, at least initially, there is a global model. In fact, it could be considered a kind of Transfer Learning in which the knowledge learned from a source domain (the weights from the whole ensemble of clients) is transferred to a target domain (each particular client). In the following, we will refer to the approach followed in this study regarding PFL as \textit{adapFL} (standing for adaptive federated learning). The idea behind this novel approach is explained in Section~\ref{sec:methodology}.

\section{Data}\label{sec:data}


\begin{sidewaysfigure}
    \centering
    \includegraphics[width=\textheight]{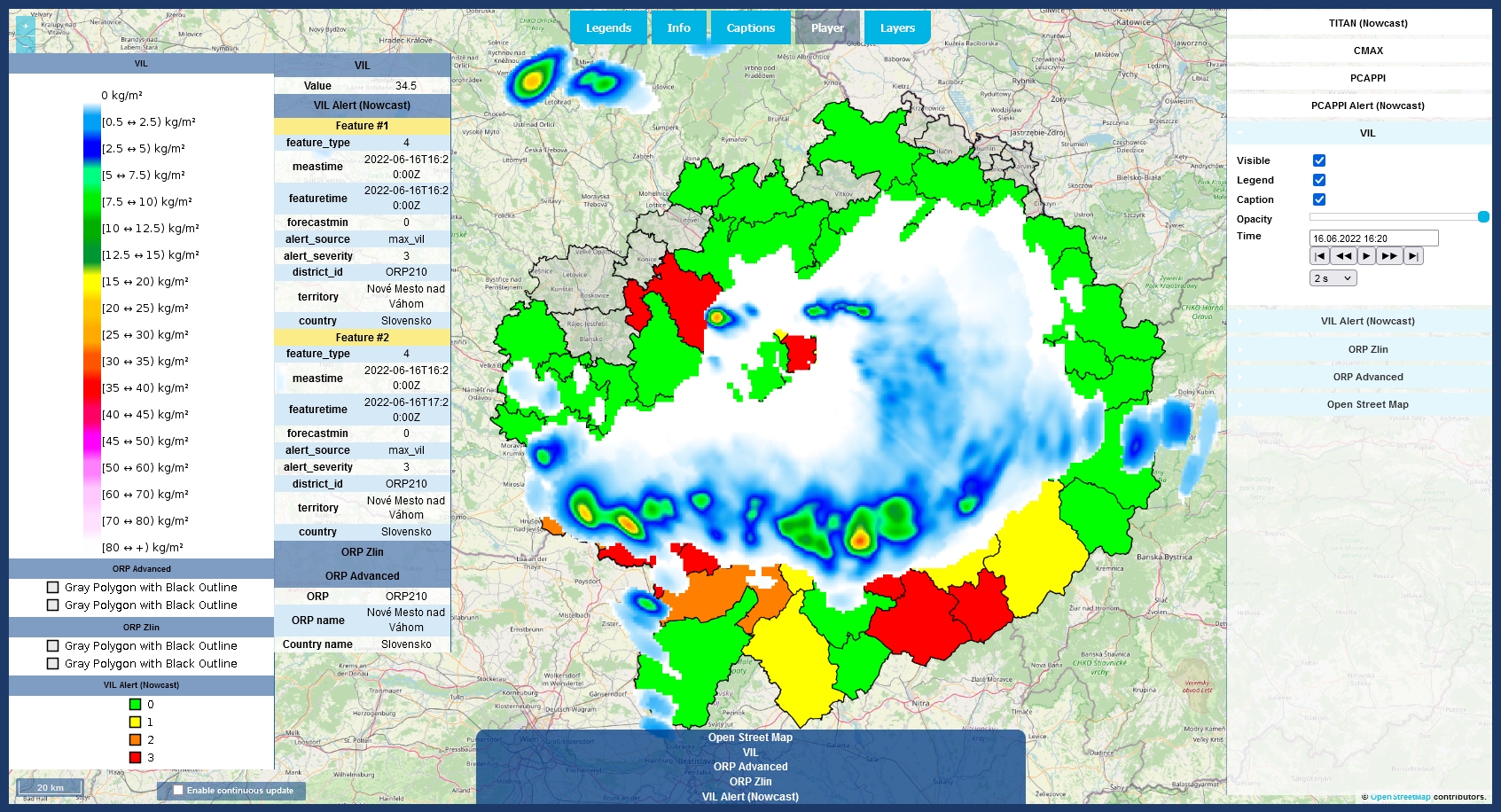} 
    \caption{Example of the area covered by the radar in the borderland between Czech and Slovak republic and the information captured: continuous white-blue-green-orange clouds - depiction of different values of radar rainfall (Vertically Integrated Liquid), green/yellow/red/orange polygons of districts - nowcast alerting regions where the rainfall is expected to be.}
    \label{fig:example_radar}
\end{sidewaysfigure}

Throughout this study, we utilize compact X-band meteorological radar, used for municipalities and in agrosector as a gap filler in large meteorological radar coverage. It is located in the borderland between Czech and Slovak republic. The radar features a parabolic antenna with a diameter 1160 mm which results in half power beam width 1.8°. The device sensitivity is 10 dBZ at 200km range and the central frequency of the X-band radar is 9410 MHz. In addition, the radiometric resolution is 8bit, the spatial resolution is 1$\times$1km and the bandwidth is 1 MHz. An example of the captured data is shown in Figure~\ref{fig:example_radar}.

The radar detects precipitation as 3D volumes of high reflectivity (radar beam reflections on water droplets, ice crystals and hail). Data are available as hdf5 files storing each day's information. In each file there are measurement products (ground truth) approximately every five minutes throughout the day and forecasts from five to forty minutes. 

There are many different types of radar products generated as various 2D sections of the 3D data, but we are focusing on vertically integrated liquid (VIL) products, i.e. reflectivity recalculated to water content by Marshall-Palmer formula and summed vertically. Then the VIL ($kg/m^{2}$) is defined as the vertical integrated quantity of liquid water content (LWC) ($kg/m^{3}$).

Radar image data related to VIL are highly dependent on the season and the time of the year. Therefore, in this case, having images captured every 5 minutes, we have chosen to use data from four months of the first available year. Specifically, we have selected the first four months available, April, May, June and July 2016, which can also represent significant periods in terms of precipitation.

\begin{figure}[ht]
    \centering
    \includegraphics[width = \linewidth]{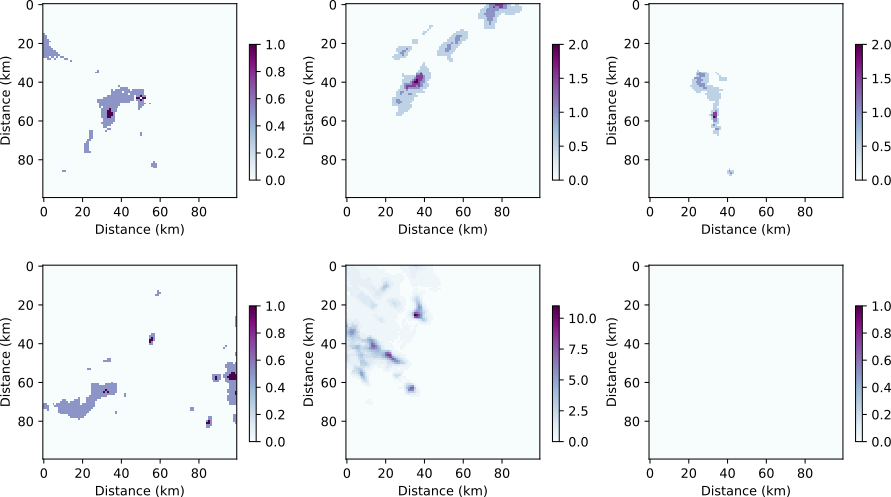}
    \caption{Example of the radar images under study after reducing to 100$\times$100 resolution with the information on vertically integrated liquid.}
    \label{fig:example_images}
\end{figure}

The original radar images are 400km wide square centered at the radar position. However, since most of the information is concentrated in the center of the images, they were cropped from 400$\times$400km to 100$\times$100km (centered in the middle of the original image). Afterwards, they were used on the different models described in Section \ref{sec:methodology} to make 5-minute predictions (nowcast) of those images. Some examples of the available images after re-scaling the resolution to the new dimensions (100$\times$100) are shown in Figure~\ref{fig:example_images}. Note that most of the information is concentrated in the center of the image due to the degradation in the radar resolution caused by the conversion of measured points coordinates from polar coordinate system to cartesian.

From Figure~\ref{fig:example_images} we can draw several conclusions. The first one is that the information in each of the images is clearly located in the center of the images. In addition, it can be seen that a large part of each image does not contain any precipitation (no rain present), most of them being totally blank images (for example, the last image in the second row). This goes hand in hand with the meteorological events that occurred at the time the radar image was acquired.

For training and testing the models, a split has been made following the temporal order of the images, so that the first 80$\%$ of images (28025) is used as the training set, and the last 20$\%$ (7007 images) is used as test set. In addition, since many of the available images are blank (no rainfall), additional processing has been performed on them, which is explained later. In addition, the validation process was performed once the data had been distributed, in order to apply the approach that should be followed in a real-world scenario. This is explained in the methodology (Section~\ref{sec:methodology}). 

Table~\ref{tab:stats_test} shows some relevant statistics according to the images to be predicted in the test set, after data processing. Specifically, these statistics are those related to the mean of the VIL in each of the images.

\begin{table}[ht]
    \centering
    \caption{Statistics regarding the mean VIL in each of the images to be predicted in the test set of each of the four zones under study. In the case of the minimum and the maximum, the average of the min and max values for each of the images are shown.}
    \begin{tabular}{cccccccc}
         \toprule
         &  \textit{Min} & \textit{Max} & \textit{Mean} & \textit{Median} & \textit{Variance} & \textit{Skewness} & \textit{Kurtosis}\\
         \midrule
         \textit{Zone 1} & 0 & 4.2192 & 0.1201 & 0.0224 & 0.0500 & 3.1100 & 12.2513\\
         \textit{Zone 2} & 0 & 3.1011 & 0.1193 & 0.0236 & 0.0592 & 4.0793 & 22.2058\\
         \textit{Zone 3} & 0 & 3.7380 & 0.1133 & 0.0249 & 0.0452 & 3.0060 & 10.0654\\
         \textit{Zone 4} & 0 & 3.2316 & 0.0956 & 0.0158 & 0.0316 & 2.8355 & 9.2811\\
         \bottomrule 
    \end{tabular}
    \label{tab:stats_test}
\end{table}

\section{Methodology} \label{sec:methodology}

In this section, the different models used to perform the 5-minutes nowcast of the VIL based on radar images are described.

\subsection{Classical models}

\textit{Tracking of Radar Echoes by Correlation} (TREC) \citep{Rinehart1978} is a nowcasting method that is based on a comparison of two consecutive images of radar reflectivity. For each block of radar pixels, TREC identifies a motion vector by maximizing the cross-correlation coefficient between the two consecutive reflectivity images. It is an image processing algorithm, which does not include any dynamic equations related to the motion and/or the evolution of the detected weather fields. 

As \citep{WooWong2017} emphasizes, ``while TREC is successful in tracking the movements of individual radar echoes, in practice, it usually captures the direction of individual rain cells instead of moving the entire meteorological system on a larger scale''. Due to this drawback of the TREC method, various alternatives such as COTREC (Continuity Tracking Radar Echoes by Correlation) \citep{NOVAK2007450}, have been developed. A recent paper \citep{Tang2018} presents a detailed overview of the features of the novel refined methods of the TREC concept that emerged during the last decades. 

In this paper, we will use COTREC as a baseline classical model to compare with the resulting deep learning models and the different learning approaches.

\subsection{Deep Learning and Federated Learning models}

As motivated in Section~\ref{sec:intro}, in view of the heterogeneous distributed nature of the radar images under study, we address this challenge with an horizontal federated learning architecture and a novel personalized federated learning method. 

\begin{figure}[ht]
    \centering
    \includegraphics[width = 0.7\linewidth]{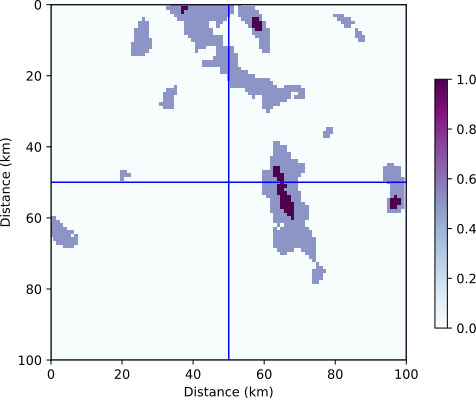}
    \caption{Example of an image where the four quadrants into which it has been divided to create the four zones are shown in blue.}
    \label{fig:split_zones}
\end{figure}

First, we divide the data artificially into 4 different zones, splitting the images into four quadrants as shown in Figure~\ref{fig:split_zones}, in which an example of a training image is shown. 

It is important to note that, for training the deep learning models, we have processed the input data in the following way: for predicting the image captured in the next 5 minutes, we introduced the 3 images available immediately before. Thus, by taking the three images we are not only be capturing the speed of the movement and the change in the amount of vertical integrated liquid, but also the acceleration. In addition, one important part of the processing carried out is that we have eliminated from both the train set and the test set those records in which the three images prior to the predictor were empty (blank, no rain). This has also been extrapolated to analyze the results with the COTREC method, taking into account only the error obtained when predicting with the remaining images. The motivation lies in the fact that in these cases, the natural answer would be to predict that there will be no rain (as the three previous images were without rain). Including in the neural network so many empty images can lead to a lack of appropriate adjustment in cases where there are heavy rain events, which as can be seen in the distribution are infrequent even after carrying out this processing (see Appendix~\ref{sec:data_distribution}), but at the same time are an important factor that nowcasting models must predict accurately. Table~\ref{tab:data_zones} shows the number of data for train and test in each of the four zones after carrying out this pre-processing.

\begin{table}[ht]
    \centering
    \caption{Number of data after processing in each zone for train and test.}
    \begin{tabular}{ccc}
         \toprule
         &  \textit{Train} & \textit{Test}\\
         \midrule
         \textit{Zone 1} & 4194 & 1585\\
         \textit{Zone 2} & 4055 & 1286\\
         \textit{Zone 3} & 4022 & 1380\\
         \textit{Zone 4} & 4457 & 1494\\
         \bottomrule 
    \end{tabular}
    \label{tab:data_zones}
\end{table}

\begin{figure}[ht]
    \centering
    \includegraphics[width = 0.7\linewidth]{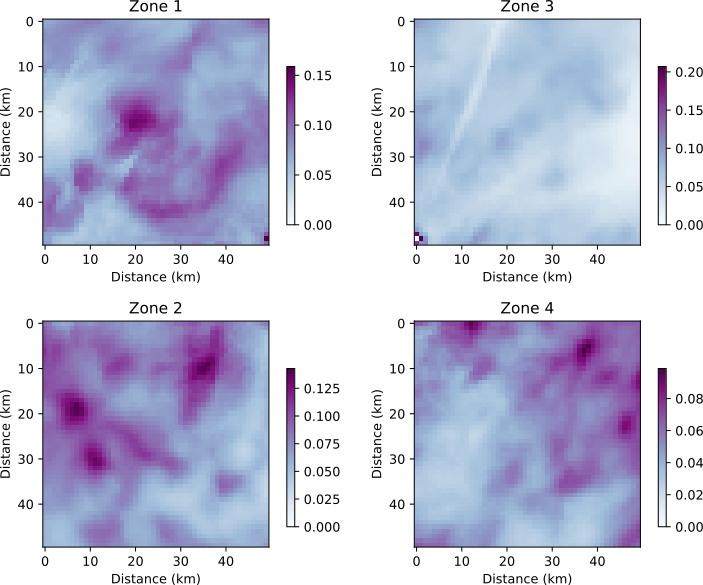}
    \caption{Average of all the radar images available in April, May, June and July 2016 by zone once processed by area.}
    \label{fig:average_images}
\end{figure}

In order to analyze the characteristics of the images available in each zone, Figure~\ref{fig:average_images} shows the mean of the images captured during the months of April, May, June and July 2016 in each artificially divided area once pre-processed as explained above. In particular, the average of the images to be predicted in train and test is shown. Note than in each case the the color bar represents the magnitude associated with the mean VIL in each pixel.

In view of Figure~\ref{fig:average_images}, we can note that the distribution of the images in each quadrant are very different from each other, being quite heterogeneous zones, as will be further explained in Section~\ref{sec:discussion}. Then, when training the model in a federated way, this may difficult the model convergence, since each model would have seen data with different statistics. However, we are interested in keeping these differences because they better reflect real use cases, where radar images will have very distinct distributions if they come from clearly differentiated areas, such as radars located in various countries. 

In addition, in order to select the model to be used for testing purposes, we need the carry out a validation process. In this case, we have performed k-fold cross validation with five folds. Time Series cross-validator from \textit{scikit-learn} has been used for this purpose in order to keep the temporal order of the images. A key point here is to decide on which dataset to perform the validation (as we would not centralize the data). Following a real use case, the four data owners (corresponding to the four zones involved) must agree on the model to be trained. In this case, in view of Table~\ref{tab:data_zones}, we have decided to carry out the validation process using the data from zone 4, as it has the largest number of data for training after processing. Several simple convolutional network architectures have been tested as well as different values for the batch size. The selected one is presented in Section~\ref{sec:cnn}.

\begin{figure}[ht]
    \centering
    \includegraphics[width=\textwidth]{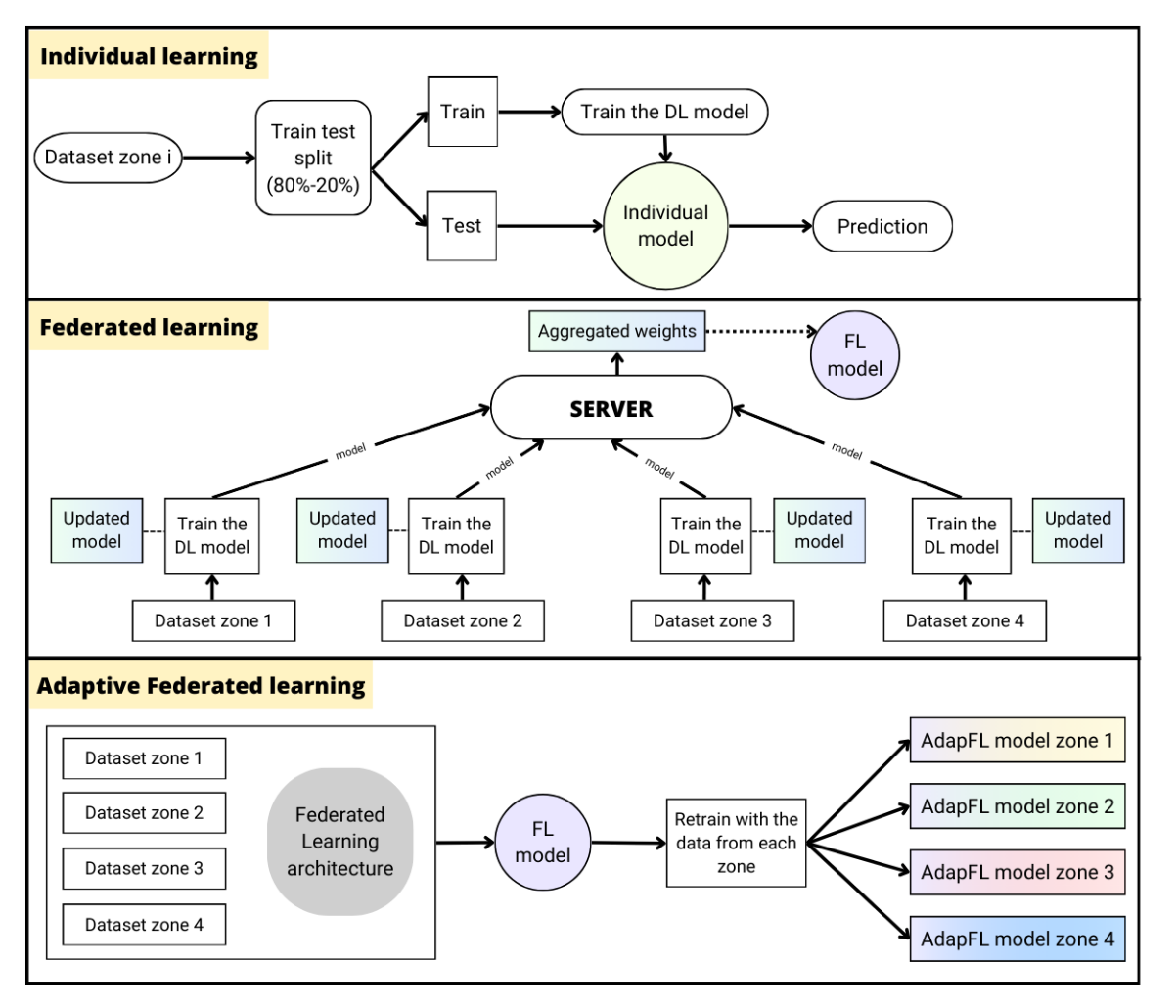}
    \caption{Schema of the three learning paradigms implemented: individual, federated and adaptive federated learning.}
    \label{fig:fl_schema_uc1}
\end{figure}

The implemented FL scheme follows the idea presented in \citep{sainzpardo2023fl}, such that the aggregation of the models trained in each zone (in this use case each zone represents a client), is performed using a weighted average according to the number of training data of each client. In addition, a novel PFL paradigm is considered once the federated training has been carried out for each client during a certain number of epochs ($N_{e}$) and a given number of rounds ($N_{r}$). In this use case, the variability in each zone with respect to the others is high, as shown in Figure~\ref{fig:example_images}. Therefore, it is appropriate that after conducting the federated training, in order to achieve a more robust model by having been trained with more data, these are trained a certain number of epochs on the data of each zone locally, which we have called \emph{adaptive federated training (\textit{adapFL})}. The idea is to be able to capture in each case the peculiarities of each zone before evaluating the results in the test set of that area, but starting with the knowledge and generalization ability provided by the FL model. In short, the three learning schemes (including the individual one) are presented in Figure~\ref{fig:fl_schema_uc1}. 

In particular, in this work we are interested in analyzing the feasibility of the \textit{adapFL} approach for the case in which images from different areas are available. The goal is to study whether this potential collaboration while maintaining the privacy of the images by not sharing them, can allow improving the results that would be obtained by training locally in each zone (as a greater number of data will be available to train the models in a distributed way), or if on the contrary, it is more convenient to carry out an individual training in each zone, respecting the differential factors of each one of them. An intuitive idea of such implemented architecture is summarized in the pseudocode given in Algorithm~\ref{alg:adapfl}. 

\begin{algorithm}[ht]
    \caption{Adaptive federated training pseudocode}\label{alg:adapfl}
    \hspace*{\algorithmicindent} \textbf{INPUT:} \textit{clients:} list with all the clients participating on the training (with $x$ the features and $y$ the labels for each client). \\
    \hspace*{\algorithmicindent} \textbf{INPUT:} \textit{n:} total number of training data from all clients.\\
    \hspace*{\algorithmicindent} \textbf{INPUT:} \textit{model:} machine learning model to be trained.\\
    \hspace*{\algorithmicindent} \textbf{INPUT:} \textit{$n_e$:} number of epochs. \\
    \hspace*{\algorithmicindent} \textbf{INPUT:} \textit{batch\_size:} batch size. \\
    \hspace*{\algorithmicindent} \textbf{INPUT:} \textit{$n_r$:} number of rounds of the FL training.
    \begin{algorithmic}[1]
    \Function{AdaptativeFederatedTraining}{$clients$, $n$, $model$, $n_e$, $batch\_size$, $n_r$}
    \For{$i \in [1,\hdots,n_r]$}
    \State $w \gets [\hspace{0.1cm}]$ \Comment{empty list for saving the weights}
    \For{$client \in clients$}
    \State \textit{model.train(client[`x'], client[`y'], epochs=$n_{e}$, batch\_size=batch\_size)}
    \State \textit{w\_client $\gets$ model.get\_weights()} 
    \State \textit{w $\gets$ w + $\left(\frac{|client[`y']|}{\sum n} \cdot w\_client\right)$} \Comment{Add to the list}
    \EndFor
    \State \textit{$w \gets \sum w$}
    \State  \textit{model.set\_weights(w)}
    \Comment{Update the model with the new aggregated weights}
    \EndFor
    \State $j \gets 1$
    \For{$client \in clients$}    \State \textit{model.train(client[`x'], client[`y'], epochs=$n_{e}$, batch\_size=batch\_size)}
    \State  \textit{model.save(f`model\_ind\_\{j\}.h5')} 
    \State $j \gets j+1$
    \EndFor
    \EndFunction
    \end{algorithmic}
\end{algorithm}

In this sense, as the federated training can be carried out in each zone in parallel, it is desirable that the total number of epochs that are trained individually ($N_{e}^{(I)}$) is equal to the number of epochs that the model is trained locally in the federated scheme ($N_{e}^{(FL)}$) times the number of rounds ($N_{r}$) of the federated training plus the number of epochs the model is trained later in each zone ($N_{e}^{(L)}$) in the \textit{adapFL} approach, i.e:

$$
N_{e}^{(I)} = N_{r}\cdot N_{e}^{(FL)} + N_{e}^{(L)}
$$

In this final implemented model the following values have been fixed: $N_{e}^{(I)} = 100$, $N_{e}^{(FL)}=N_{e}^{(L)}=10$ and $N_{r}=10$ for the classic FL approach and $N_{r}=9$ for the \textit{adapFL} approach.

\subsection{Convolutional neural architecture}\label{sec:cnn}
Both in the case of individual learning in each zone and in the case of the \textit{adapFL} schema, we have developed a neural network architecture composed of the following convolutional layers presented in Table~\ref{tab:cnn} (further detailed in  Figures~\ref{fig:ANN_uc1} and \ref{fig:ANN_uc1_2} from Appendix~\ref{sec:app_cnn}):


\begin{table}[ht]
\caption{Convolutional neural network implemented, layers, output shape, number of parameters and activation function applied.}
\label{tab:cnn}
\begin{tabular}{llll}
\hline
\multicolumn{1}{c}{\textbf{Layer (type)}} & \multicolumn{1}{c}{\textbf{Output Shape}} & \multicolumn{1}{c}{\textbf{Param \#}} & \textbf{Activation} \\ \hline
conv2d\_46 (Conv2D)                         & (None, 48, 48, 128)                        & 3584                                   & Relu       \\ 
conv2d\_47 (Conv2D)                         & (None, 46, 46, 64)                         & 73792                                  & Relu       \\ 
conv2d\_48 (Conv2D)                         & (None, 44, 44, 32)                         & 18464                                  & Relu       \\ 
conv2d\_49 (Conv2D)                         & (None, 42, 42, 16)                         & 4624                                   & Relu       \\ 
conv2d\_50 (Conv2D)                         & (None, 40, 40, 16)                         & 2320                                   & Relu       \\ 
conv2d\_51 (Conv2D)                         & (None, 38, 38, 32)                         & 4640                                   & Relu       \\ 
conv2d\_52 (Conv2D)                         & (None, 36, 36, 64)                         & 18496                                  & Relu       \\ 
conv2d\_53 (Conv2D)                         & (None, 34, 34, 128)                        & 73856                                  & Relu       \\ 
conv2d\_54 (Conv2D)                         & (None, 32, 32, 1)                          & 1153                                   & Relu       \\ \hline
\end{tabular}
\end{table}

The architecture is kept purposely simple, as the intent of this paper is to evaluate the potential of the FL and PFL approaches for this use case, not to establish a new SotA model in precipitation nowcasting based on radar images. 

The model has been compiled with \textit{Adam} \citep{kingma2014adam} as optimizer and the \textit{mean squared error (MSE)} as loss function and the MSE and the \textit{mean absolute error (MAE)} as quality metrics for monitoring.

Again, as previously explained, this model has been validated using the data from zone 4. Specifically, the average MSE obtained in the five folds is 0.0326. 

In addition, we have calculated other metrics such us the RMSE and the \textit{skill score} (like in \cite{KUMAR2024103600}), which can be obtained from the MSE, so the tendency observed will be analogous, but they can be used for comparison with other cases. The RMSE is defined as the squared root of the mean of the squared errors, and the skill score for the model $\mathcal{M}$ ($SkillScore_{\mathcal{M}}$ can be defined as given in Equation~\ref{eq:skill_score}, being $MSE_{\mathcal{M}}$ the MSE for the model $\mathcal{M}$ and $MSE_{\mathcal{B}}$ the MSE for a baseline model. In our study we have considered COTREC as the baseline model for calculating the skill score.

\begin{equation}\label{eq:skill_score}
    SkillScore_{\mathcal{M}} = 1 - \frac{MSE_{\mathcal{M}}}{MSE_{\mathcal{B}}}
\end{equation}

\subsection{Computing resources}\label{sec:computing}

The experiments have been carried out using the AI4EOSC platform \citep{ai4eosc}. Specifically, an environment with 16 GB of RAM, 10 GB of disk, 8 CPU cores, and one GPU NVIDIA Tesla T4 has been deployed. The programming language used was Python 3 and the TensorFlow library \citep{tensorflow2015-whitepaper} was used to develop the deep learning models.

\section{Results and analysis}\label{sec:results}

In this section, the results obtained in each of the zones when evaluating each of the following approaches are presented:

\begin{itemize}
    \item The classic COTREC model as baseline,
    \item An individual learning model trained for 100 epochs in the same zone (IL),
    \item An FL model trained 10 epochs for 10 rounds (FL), 
    \item A PFL model trained 10 epochs for 9 rounds, ending with a local training of 10 epochs on the data of the testing zone (\textit{adapFL}).
\end{itemize}

Table~\ref{tab:mse_mae_zones} summarizes the results obtained in terms of the MSE and the MAE in the test set of each of the four zones and with the four approaches.


In view of the results shown in Table~\ref{tab:mse_mae_zones}, the best results are obtained with the \textit{adapFL} architecture. Moreover, all the three DL approaches achieve better results than the COTREC model. It should be noted that with the conventional FL architecture, the results of the COTREC model were improved substantially, but not those of the IL, since the latter has a better capacity to adapt to the data captured in its corresponding area. Thus, with the adaptive step of the \textit{adapFL} architecture, we managed to improve the results of the FL architecture, but also those of the IL approach. Note that the adaptive step in \textit{adapFL} can be seen as a kind of transfer learning from the FL model. In Figures~\ref{fig:adapfl_pred} and \ref{fig:adapfl_pred2} of Appendix~\ref{sec:app_pred}, it can be observed for each of the four zones, six examples of observed images together with the corresponding prediction obtained with the best analyzed model according to the results presented in Table~\ref{tab:mse_mae_zones}, which corresponds to the \textit{adapFL} approach.
Within those figures we can observe that the predictions obtained are very similar to the actual observed images in the six cases shown for each of the four zones. 

Note that we have calculated these metrics, both MSE and MAE, taking the mean error in each image, and then the mean of all of them as stated in \cite{han2023key} and \cite{jolliffe2012forecast}. 


\begin{table}[htbp]
    \centering
    \caption{MSE and MAE ($kg/m^{2}$) obtained in the test set with the different approaches in each zone.}
    \label{tab:mse_mae_zones}
        \begin{tabular}{cccccc c ccc}
             \toprule
             & \multicolumn{4}{c}{\textit{MSE}} & & \multicolumn{4}{c}{\textit{MAE}}\\
             \cmidrule{2-5}\cmidrule{7-10}
             \textit{Zone} &  \textit{COTREC} & \textit{IL} & \textit{FL} & \textit{adapFL} &  &\textit{COTREC} & \textit{IL} & \textit{FL} & \textit{adapFL}\\
             \midrule
             \textit{1} & 0.2306 & 0.1399 & 0.2000 & \textbf{0.1352} & & 0.0708 & 0.0618 & 0.0649 & \textbf{0.0606}\\
             \textit{2} & 0.1756 & 0.1057 & 0.1612 & \textbf{0.1005} & & 0.0650 & 0.0530 & 0.0590 & \textbf{0.0517}\\
             \textit{3} & 0.2235 & 0.1220 & 0.1286 & \textbf{0.1082} & & 0.0729 & 0.0607 & 0.0576 & \textbf{0.0565}\\
             \textit{ 4} & 0.1095 & 0.0852 & 0.0958 & \textbf{0.0815} & & 0.0514 & 0.0482 & 0.0484 & \textbf{0.0464}\\
             \bottomrule 
        \end{tabular}
\end{table}

The idea of taking the current configuration of the \textit{adapFL} approach is motivated by achieving an adaptation by customizing the conditions of each specific zone, while trying to preserve the generalization capacity of the FL architecture. However, we have sought to compare the different configurations, given by \textit{($N_{r}$, $N_{e}^{(L)}$)}, $\forall N_{r}\in\{1,\hdots,10\}$, and $N_e^{(L)}=100-(10\cdot N_r)$, with $N_r$ the number of rounds of the FL architecture and $N_e^{(L)}$ the number of epochs that the model has been trained specifically in each zone after obtaining the federated global model (note that if $N_{r}=10$, we are in the case of the classic FL approach). The results as a function of the MSE and the MAE for each test set are displayed in Figures~\ref{fig:mse_test_adapFL_il_fl} and \ref{fig:mae_test_adapFL_il_fl} from Appendix~\ref{sec:app_mse_mae}. It can be observed that that a lower value of $N_r$ is related to a reduction of the MAE and the MSE in train, which is explained as a higher adjustment to the zone under analysis. However, this is also linked to a reduction of the generalization capability that can be extracted from the federated architecture. Therefore, the selected configuration is (9, 10), i.e. 9 rounds of the FL architecture and 10 final epochs of personalized training in each zone.  

Finally, the RMSE and skill score for the four areas can be calculated from the MSE pressented in Table~\ref{tab:mse_mae_zones}, and are shown in Table~\ref{tab:rmse_skill_score}. In view of the skill score from Table~\ref{tab:rmse_skill_score} we can note that the greatest improvement of the best approach (in this case \textit{adapFL}) in relation to the baseline model (COTREC), is obtained for zone 3. As the objective of the skill score is to observe whether the model is better than a given base model, we have highlighted for each zone the greater one (which is \textit{adapFL} in all the four cases).

\begin{table}[htbp]
    \centering
    \caption{RMSE ($kg/m^{2}$) and skill score obtained in the test set with the different approaches in each zone.}
    \label{tab:rmse_skill_score}
        \begin{tabular}{cccccc c ccc}
             \toprule
             & \multicolumn{4}{c}{\textit{RMSE}} & & \multicolumn{4}{c}{\textit{Skill Score}}\\
             \cmidrule{2-5}\cmidrule{7-10}
             \textit{Zone} &  \textit{COTREC} & \textit{IL} & \textit{FL} & \textit{adapFL} &  &\textit{COTREC} & \textit{IL} & \textit{FL} & \textit{adapFL}\\
             \midrule
             \textit{1} & 0.4802 & 0.3740 & 0.4472 & \textbf{0.3677} & & - & 0.3933 & 0.1327 & \textbf{0.4137} \\
             \textit{2} & 0.4190 & 0.3251 & 0.4015 & \textbf{0.3170} & & - & 0.3981 & 0.0820 & \textbf{0.4277} \\
             \textit{3} & 0.4728 & 0.3493 & 0.3586 & \textbf{0.3289} & & - & 0.4541 & 0.4246 & \textbf{0.5159} \\
             \textit{4} & 0.3309 & 0.2919 & 0.3095 & \textbf{0.2855} & & - & 0.2219 & 0.1251 & \textbf{0.2557} \\
             \bottomrule 
        \end{tabular}
\end{table}

Finally, the distribution of the MAE and the MSE for each image of the test set of each zone is shown in Appendix~\ref{sec:app_mse_mae_images} by means of histograms. 

\subsection{Central area}
\begin{figure}[ht]
    \centering
    \includegraphics[width = 0.7\linewidth]{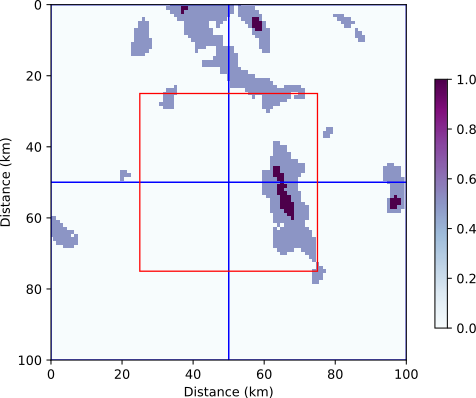}
    \caption{Example of an image where the four quadrants into which it has been divided to create the four zones are shown in blue and the central area is marked in red.}
    \label{fig:central_zone}
\end{figure}

Once we have the four individual models and the models trained in a federated way, we are interested in seeing how the last ones perform in a different region. To do this, we take the 50$\times$50 central crop of the original image (so that it is of the same resolution as the images of each of the 4 zones). An example of an image where this division is taken is shown in Figure~\ref{fig:central_zone}, where the central area selected is the one framed within the red square.

Let us consider all the images within this quadrant and evaluate the predictions that would be obtained with the FL model and the four adaptive FL models for each of the 4 initial areas. Also, we compare it with the IL model on that area and the \textit{adapFL} one with the adaptive phase performed in such zone (the central one), but the FL model only on the first 4 zones. Since part of the images that we evaluate are part of the training set of each individual zone, we will take the same test set as in the previous cases, corresponding to the last 20$\%$. Again, the same processing will be carried out in each zone regarding the blank (empty, no-rain) images and the input for the DL models. The results obtained in each case for the train and test set and both for the MSE and the MAE are shown in Table~\ref{tab:mse_central_zone}. Note from this table that the first three rows correspond to models that have explicitly seen data from the central zone training set (COTREC, IL over the central zone, and \textit{adapFL} with FL over the four initial zones and personalized with the adaptive phase over the central area with a configuration (9,10)). The rest of the models shown in Table~\ref{tab:mse_central_zone} have not directly seen training data from the central zone, although each of the four initial zones contain part of the data from the central area.

\begin{table}[ht]
    \centering
    \caption{Comparison of the MSE and MAE ($kg/m^{2}$) obtained in the train and test sets of the central zone with the different approaches analyzed.}
    \begin{tabular}{rcccc}
    \toprule
         & \multicolumn{4}{c}{\textit{Central zone}}\\
         \cmidrule{2-5}
         & \textit{MSE} & \textit{MSE} & \textit{MAE} & \textit{MAE}\\
         \textit{Model} & \textit{Train} & \textit{Test} & \textit{Train} & \textit{Test}\\
         \midrule
         \textit{COTREC} & 0.0798 & 0.2547 & 0.0308 & 0.0735\\ 
         \textit{IL zone central} & 0.0118 & \textbf{0.0121} & \textbf{0.0130} & 0.0207\\
         \textit{adapFL (central)} & \textbf{0.0067} & 0.0129 & 0.0149 & \textbf{0.0199}\\
         \midrule
         \textit{FL (four areas)} & 0.0128 & 0.0140 & \textbf{0.0161} & 0.0195\\
         \textit{adapFL (zone 1)} & 0.0111 & 0.0130 & 0.0174 & 0.0210\\
         \textit{adapFL (zone 2)} & \textbf{0.0103} & 0.0134 & 0.0165 & 0.0201\\
         \textit{adapFL (zone 3)} & 0.0112 & \textbf{0.0123} & 0.0169 & 0.0200\\
         \textit{adapFL (zone 4)} & 0.0114 & \textbf{0.0123} & 0.0159 & \textbf{0.0187}\\
         \bottomrule 
    \end{tabular}
    \label{tab:mse_central_zone}
\end{table}

From Table~\ref{tab:mse_central_zone}, we can highlight the following points: (1) it makes sense to analyze the performance of the FL model since each initial area has part of the central zone analyzed in this case. (2) Intuitively, the best results should be obtained with the models trained in the same zone, in this case IL in the central zone and \textit{adapFL} on the central zone. (3) The COTREC model is largely worse than the other approaches analyzed, so it can be extrapolated that in this scenario it is more convenient to apply DL models. (4) In the training set the \textit{adapFL} model over the central zone is the best in terms of MSE, and the individual one with respect to the MAE (followed by \textit{adapFL} over the central zone). (5) In the test set the best results for MSE are obtained with the individual model, followed by the \textit{adapFL} ones over zones 3 and 4. (6) In the test set the best result with respect to the MAE is obtained with the \textit{adapFL} model over zone 4 followed by the basic FL model, being in this case the FL approach better than the individual model trained over the current area of study. 

Again in this case we have calculated from Table~\ref{tab:mse_central_zone} the RMSE and the skill score (with COTREC as baseline) for the prediction on the central zone. The results for these two metrics can be found in Table~\ref{tab:rmse_skillscore_zone}. Note that in this case the results for the skill score show a greater difference between the MSE for the baseline model and the other approaches than in the previous case where the four initial zones were analyzed.

\begin{table}[ht]
    \centering
    \caption{Comparison of the MSE ($kg/m^{2}$) and skill score obtained in the train and test sets of the central zone with the different approaches analyzed.}
    \begin{tabular}{rcccc}
    \toprule
         & \multicolumn{4}{c}{\textit{Central zone}}\\
         \cmidrule{2-5}
         & \textit{RMSE} & \textit{RMSE} & \textit{Skill Score} & \textit{Skill Score}\\
         \textit{Model} & \textit{Train} & \textit{Test} & \textit{Train} & \textit{Test}\\
         \midrule
         \textit{COTREC} & 0.2825 & 0.5047 & - & -\\ 
         \textit{IL zone central} & 0.1086 & \textbf{0.1100} & 0.8521 & \textbf{0.9525}\\ 
         \textit{adapFL (central)} & \textbf{0.0819} & 0.1136 & \textbf{0.9160} & 0.9494 \\ 
         \midrule
         \textit{FL (four areas)} & 0.1131 & 0.1183 & 0.8396 & 0.9450\\ 
         \textit{adapFL (zone 1)} & 0.1054 & 0.1140 & 0.8609 & 0.9490\\ 
         \textit{adapFL (zone 2)} & \textbf{0.1015} & 0.1158 & \textbf{0.8709} & 0.9474\\ 
         \textit{adapFL (zone 3)} & 0.1058 & \textbf{0.1109} & 0.8596 & \textbf{0.9517}\\ 
         \textit{adapFL (zone 4)} & 0.1068 & \textbf{0.1109} & 0.8571 & \textbf{0.9517}\\
         \bottomrule 
    \end{tabular}
    \label{tab:rmse_skillscore_zone}
\end{table}

\begin{figure}[ht]
    \centering
    \includegraphics[width=\linewidth]{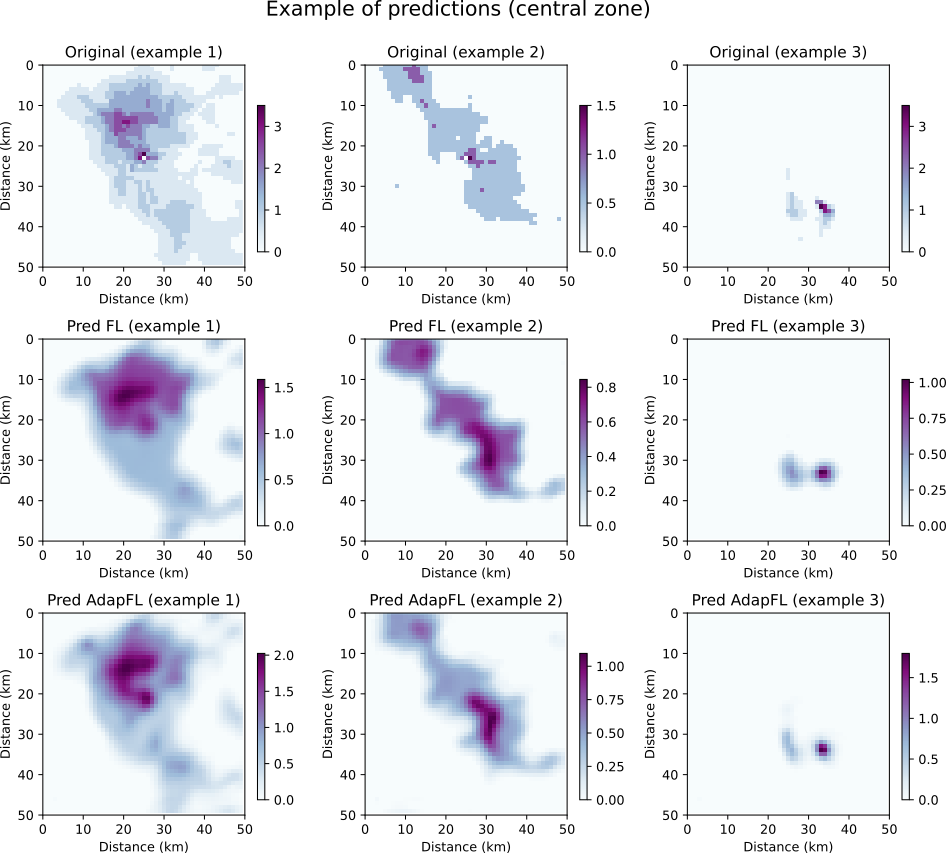}
    \caption{Example of predictions of the test set of the central zone. Federated learning approach with $N_{r}=10$ and $N_{e}=10$ and \textit{adapFL} method with the configuration (9,10).}
    \label{fig:central_pred}
\end{figure}

Furthermore, Figure~\ref{fig:central_pred} shows three images in the conditions mentioned above (central quadrant of resolution $50\times50$), and their corresponding prediction obtained using the FL approach applied to the four initial areas and the \textit{adapFL} one (adapted to the central area).

Finally, it is interesting to analyze the non-independent identically distributed (i.i.d.) nature of each of the five zones proposed. Assuming the privacy-preserving deep learning approach (FL and \textit{adapFL}) in which we do not have access to all the raw images in order to preserve privacy, technical or legal considerations that may apply to each image (e.g. may belong to different data owners), this analysis is carried out on the only information that the central server that performs the federated training would have available: the weights that define the model. In this sense, the divergence of the weights between each pair of zones \citep{fl_noniid} was analyzed. Specifically, we repeated the training in the 5 zones (including the central zone) from the initial model when training only 10 epochs individually (as done in each round of the FL training), and we kept the weights obtained in each case. Be $||\cdot||_{F}$ the Frobenius norm, and $w_{i}^{(n)}$ the weights obtained when training with the data from client $i$ in round $n$, we define the divergence in a symmetric way $d_{i,j}^{(n)}$ as follows:

\begin{equation}
    d_{i,j}^{(n)}=\frac{||w_{i}^{(n)}-w_{j}^{(n)}||_{F}}{\frac{1}{2}(||w_{i}^{(n)}||_{F}+||w_{j}^{(n)}||_{F})}
    \label{eq:divergence}
\end{equation}

Equation~\ref{eq:divergence} presents a modification regarding the divergence equation proposed in \cite{fl_noniid} in order to make it symmetric, In order to do so the norm of the differences is divided by the mean of the Frobenius norm of each weights. 

Specifically, in order to calculate the norm of the weights corresponding to each grid, the corresponding norm of each layer has been calculated individually and then the norm of the resulting vector has been computed. The results obtained in each case for $d_{i,j}^{(1)}$ are summarized in Table~\ref{tab:divergences}. 

\begin{table}[ht]
    \centering
    \caption{$d_{i,j}$ calculated for each combination of zones.}
    \begin{tabular}{r|ccccc}
       \textit{i,j}  & \textit{1} & \textit{2} & \textit{3} & \textit{4} & \textit{Central} \\
       \midrule
       \textit{1} & - & 0.6563 & 0.6310 & 0.6413 & 0.5729 \\
       \textit{2} & 0.6563 & - & 0.6275 & 0.6296 & 0.5850 \\
       \textit{3} & 0.6310 & 0.6275 & - & 0.6055 & 0.5677 \\
       \textit{4} & 0.6413 & 0.6296 & 0.6055 & - & 0.5743 \\
       \textit{Central} & 0.5729 & 0.5850 & 0.5677 & 0.5743 & - \\
    \end{tabular}
    \label{tab:divergences}
\end{table}

In view of the values obtained for $d_{i,central}$ $\forall i \in \{1,2,3,4\}$, the lowest value is obtained with respect to the third zone, which is consistent with this being the \textit{adapFL} model that gives the best results in the central zone in the test set regarding the MSE. However, the best test result is obtained with zone 4 (for MAE and 3 and 4 for the MSE), which is the second with the greatest divergence, and is also the model with the greatest error in the train (of the indexed ones). This high divergence in training is reflected in the MSE of the train, while allowing a better generalization in the test. Presumably, this is due to the rainfall levels in zone 4 during the period incorporated in the training, which will be similar to that of the central zone test (covering summer periods). It is evident that the divergence between the central zone and the other four areas represents the lowest values (last row of Table~\ref{tab:divergences}), since each of the 4 initial zones contains a quarter of the information of the central one. The table above provides information on the differences between the different zones, highlighting for example that the smallest divergence regarding the initial four areas is reached between zones 3 and 4, so it may be reasonable to create models agreed between these two areas, separately from the other two, making clients' clusters. Finally, note that zone 1 presents the greater divergence with respect to the other three initial areas, so including this one may be damaging the performance of the overall FL model.

\section{Discussion}\label{sec:discussion}

In this work, we have analyzed the applicability of a novel PFL approach in comparison with a DL model trained individually on each set of data and with a classic FL architecture, for precipitation nowcasting based on radar images. However, it is important to discuss the limitations of the data used in this study as a benchmark to extrapolate the applicability of the \textit{adapFL} architecture to use cases with a sufficient number of available clients.

The first limitation is that we have information captured by a single radar. However, the coverage area of this radar is so wide and representative that it allows us to carry out an artificial division in such a way that we distribute the captured images into different clients, each of them composed of different parts of the original images. It is important to note that in a real use case scenario in which we would have radars that capture images with different resolutions, it would be necessary to perform a careful preprocessing. In the same way, in this case, we have the advantage of all clients having images captured with the same type of radar. This is important to take into account because in a case where we have different radars, the sensitivity of these as well as their measurement parameters are determining factors when it comes to damage the accuracy of a jointly built deep learning model.

Another particularity of the data used in this study is that the information contained in the images usually lies in the center of the initial full image. Therefore, dividing them in four quadrants to have four artificial clients results in a high divergence among them. In consequence, while the client with the data from the upper left quadrant of the original image will have more information in the lower right corner, this is reversed for the client with the images in the lower right quadrant. However, this is convenient in order to compare this case and use it as a benchmark to extrapolate to different scenarios with multiple radars, since in both cases such divergences will also be appreciated, being data captured in different regions or countries.

In addition, as shown in Figure~\ref{fig:example_images}, many parts of the images do not contain relevant precipitation events, besides being data with a high seasonal component. This characteristic of the data would have added more complexity during model training and evaluation so the data have been pre-processed accordingly.

These limitations are important to understand the scope and constraints of our study. Nonetheless, in spite of the advantages and disadvantages of using a single client to simulate the presence of four clients to perform FL, the obtained results are robust enough to assess the viability of the novel PFL approach 
\textit{adapFL}. As revealed by the skill score, this novel approach had a better performance in the four zones under study, in comparison with the IL approach and the classical FL one. To the best of our knowledge, it is the first time that the FL paradigm is applied to enhance the precipitation nowcasting using radar images and this contribution may be of interest to other researchers and practitioners that develop DL models for such purpose, including other convolutional ANN architectures and even with the addition of more input variables, as proposed by \cite{KUMAR2024103600}. This work has demonstrated that the use of FL and PFL techniques in regression tasks not only presents the inherent advantages of this methods, which are mentioned in the introduction, but also it is possible to reach better predictions in terms of error. Therefore, even in the case that no privacy issues affect the use of the radar images, the \textit{adapFL} approach is of particular interest. 

\section{Conclusions}\label{sec:conclusions}

Our main objective towards this study was to analyze the feasibility of a novel PFL architecture, in particular the so-called \textit{adapFL} approach, to a use case of meteorological radar images. Specifically, the objective was to use the available images to predict the precipitation expressed as VIL in the next 5 minutes. The complexity of this task, involving the use of images, invites using DL models based on convolutional neural networks.

In particular, to analyze the applicability of the PFL technique to this case, the results obtained in the four artificially distributed zones were compared with the COTREC method, as well as with the training of the same convolutional network on each of the zones individually and the training with a conventional FL architecture. Specifically, for the test set, we obtained improvements in all four cases when applying the \textit{adapFL} architecture both for the MSE and MAE. Moreover, when we extrapolated these results to the central zone, which has a part in each of the four initial ones, the optimal results for the training set regarding the MSE are also achieved with the application of the \textit{adapFL} architecture trained on all the four initial areas and adapted to the central one. Regarding only the individual models, in this case, the best results are achieved with the individual training on the fourth zone, which is in contrast with the results obtained for the divergence between the different artificial regions. This may be attributed to these regions having more in common in the test set than in the train, e.g. due to seasonality.

The results obtained throughout this study give us a promising idea about the applicability of this PFL architecture to this type of meteorological radar images, since as mentioned, the results improve those of individual training in all the analyzed cases and that of the classic COTREC method. In this sense, future work has been drawn from the very approach of this benchmark study, and it is to extrapolate this type of analysis to radar data distributed in different countries or regions. In that case, the differential behavior of each zone may lead to the introduction of measures considering these divergences, like the creation of different clusters of clients. It should be noted that in this study one of the zones was very different regarding the data used for training from the others on average (see Figure~\ref{fig:average_images}, zone 3), but the model was not degraded. Instead, this variability in the data distribution in the different clients in some cases may even lead to better generalization ability in the event of unseen data.
\\

\backmatter





\bmhead{Acknowledgements}

The authors acknowledge the funding and the support from the project AI4EOSC ``Artificial Intelligence for the European Open Science Cloud'' that has received funding from the European Union's Horizon Europe research and innovation programme under grant agreement number 101058593.

\section*{Declarations}



\bmhead{Funding}

This work has been funded by the project AI4EOSC ``Artificial Intelligence for the European Open Science Cloud'' that has received funding from the European Union's Horizon Europe research and innovation programme under grant agreement number 101058593.

\bmhead{Declaration of Competing Interest}

The authors declare that they have no known competing financial interests or personal relationships that could have appeared to influence the work reported in this paper.

\bmhead{Author Contributions Statement}



J.S-P.D. conducted the research, performed the experiments and the training of the models and wrote the main manuscript. M.C., J.B., I.H.C., K.A., L.B. and A.L.G. contributed to writing-reviewing the manuscript.  
J.B., I.M.O. and I.M. acquired, processed, curated and stored the original data used in this study.  
J.S-P.D., M.C., J.B., I.H.C., K.A., L.B, V.K. and A.L.G. participated in scientific discussions that led to this research.
V.K. and A.L.G. contributed to project management and funding acquisition.







\newpage
\begin{appendices}

\section{Data distribution}\label{sec:data_distribution}

\begin{figure}[ht]
    \centering
    \includegraphics[width=0.9\textwidth]{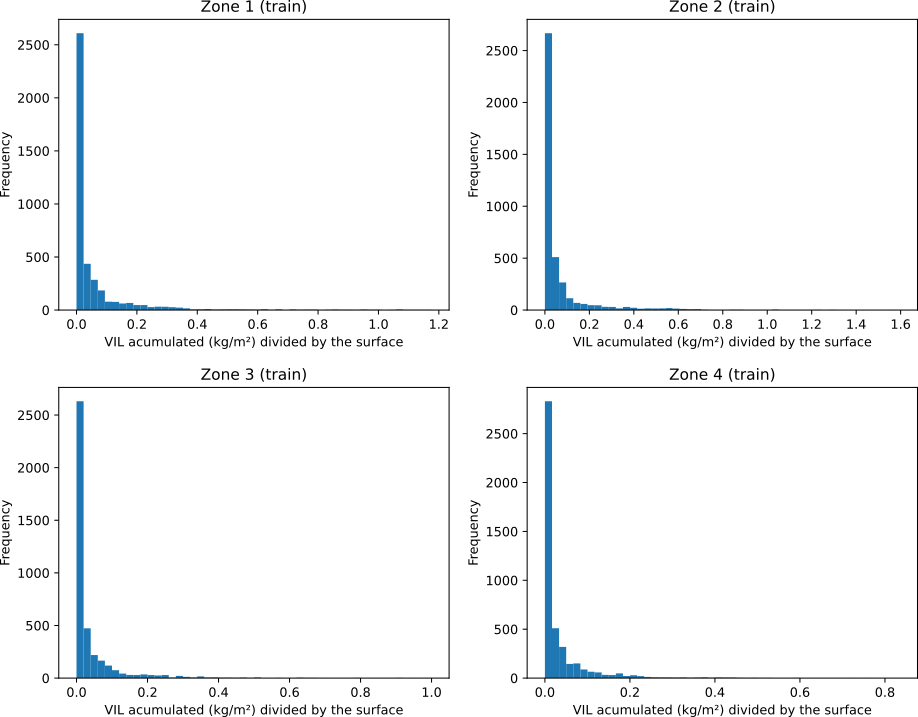}
    \caption{Distribution of the VIL accumulated in the train data (y) in each zone after processing.} 
    \label{fig:dist_train}
\end{figure}

\begin{figure}[ht]
    \centering
    \includegraphics[width=0.9\textwidth]{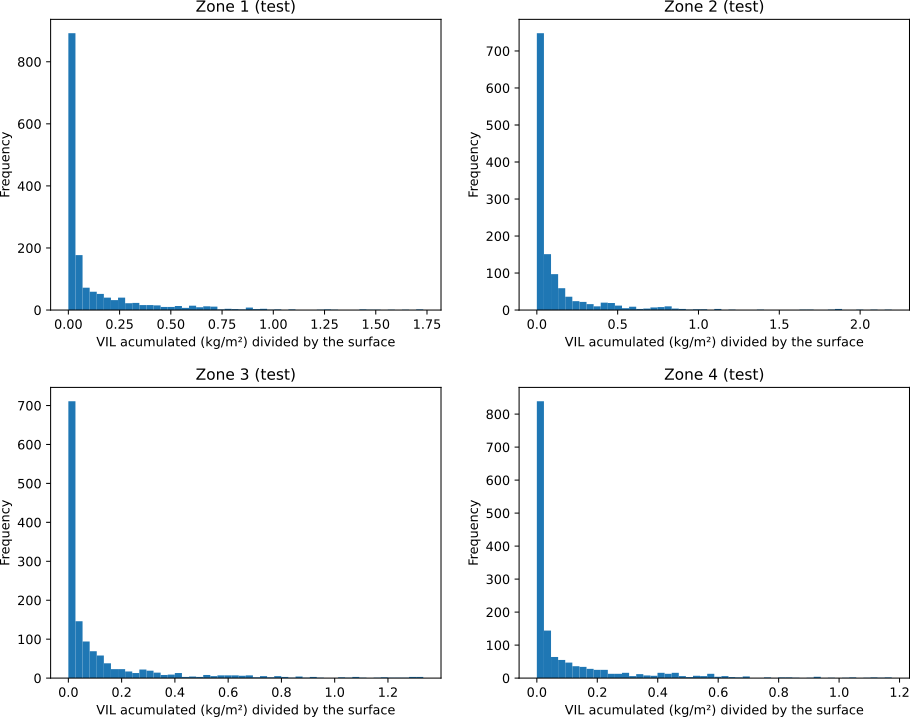}
    \caption{Distribution of the VIL accumulated in the test data (y) in each zone after processing.}
    \label{fig:dist_test}
\end{figure}

As previously stated, the main objective of this work is to predict rainfall events (given by the VIL) 5 minutes in advance. The deep learning models used for this purpose include as input the three available images prior to the time of the prediction. In order to avoid over-training the models with non-rain events, we have eliminated those cases where the predictand (the three images previously available at the time of prediction), are empty (blank) images (no rain). Eliminating these data significantly reduces the amount of training data. This processing has been carried out individually on the train and test sets of each of the zones, as would be appropriate in cases where the data are distributed. Thus, Figure~\ref{fig:dist_train} shows the accumulated VIL in each image to be predicted in the train in each of the zones, and in Figure~\ref{fig:dist_test}, the corresponding VIL in the test set in each case. To calculate the accumulated VIL displayed in the following images we have summed the VIL associated with each pixel of a given image and divided it by the surface of the image ($50^{2}$ $km^{2}$).


\clearpage
\section{\textit{AdapFL} architecture: MSE and MAE}
\label{sec:app_mse_mae}
\begin{figure}[ht]
    \centering
    \includegraphics[width=0.75\linewidth]{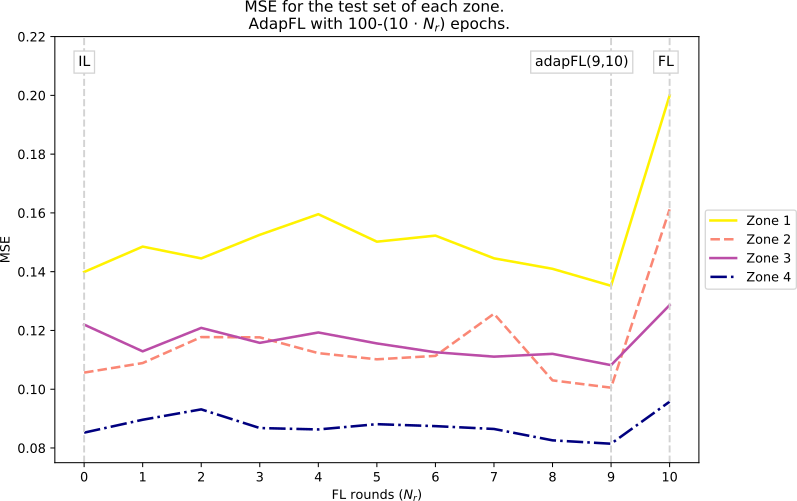}
    \caption{MSE for the test set of each zone under different \textit{adapFL} configurations, including IL and FL.}
    \label{fig:mse_test_adapFL_il_fl}
\end{figure}

\begin{figure}[ht]
    \centering
    \includegraphics[width=0.75\linewidth]{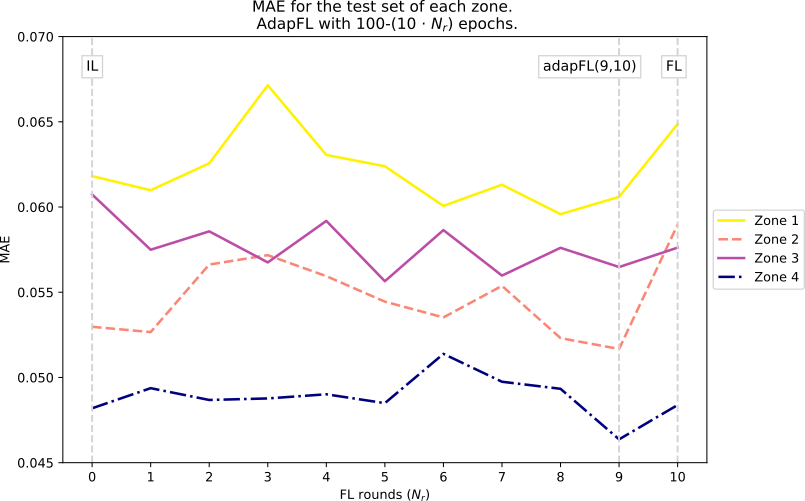}
    \caption{MAE for the test set of each zone under different \textit{adapFL} configurations, including IL and FL.}
    \label{fig:mae_test_adapFL_il_fl}
\end{figure}

In this section we aim to compare the different configurations of the proposed \textit{adapFL} architecture. Specifically, this configurations are given by \textit{($N_{r}$, $N_{e}^{(L)}$)}, $\forall N_{r}\in\{1,\hdots,10\}$, and $N_e^{(L)}=100-(10\cdot N_r)$, as explained in Section~\ref{sec:results}. Figures~\ref{fig:mse_test_adapFL_il_fl} and \ref{fig:mae_test_adapFL_il_fl} show the MSE and the MAE for the test set of each zone with each proposed configuration of the \textit{adapFL} approach. Note than in the case in which $N_{r}=0$, we are displaying the results for the individual approach (IL), and when $N_{r}=10$, those of the classic FL approach.

\vspace{-1cm}

\clearpage
\section{Example of predictions obtained for each area}
\label{sec:app_pred}
\begin{figure}[ht!]
    \centering
    \includegraphics[width=0.68\linewidth]{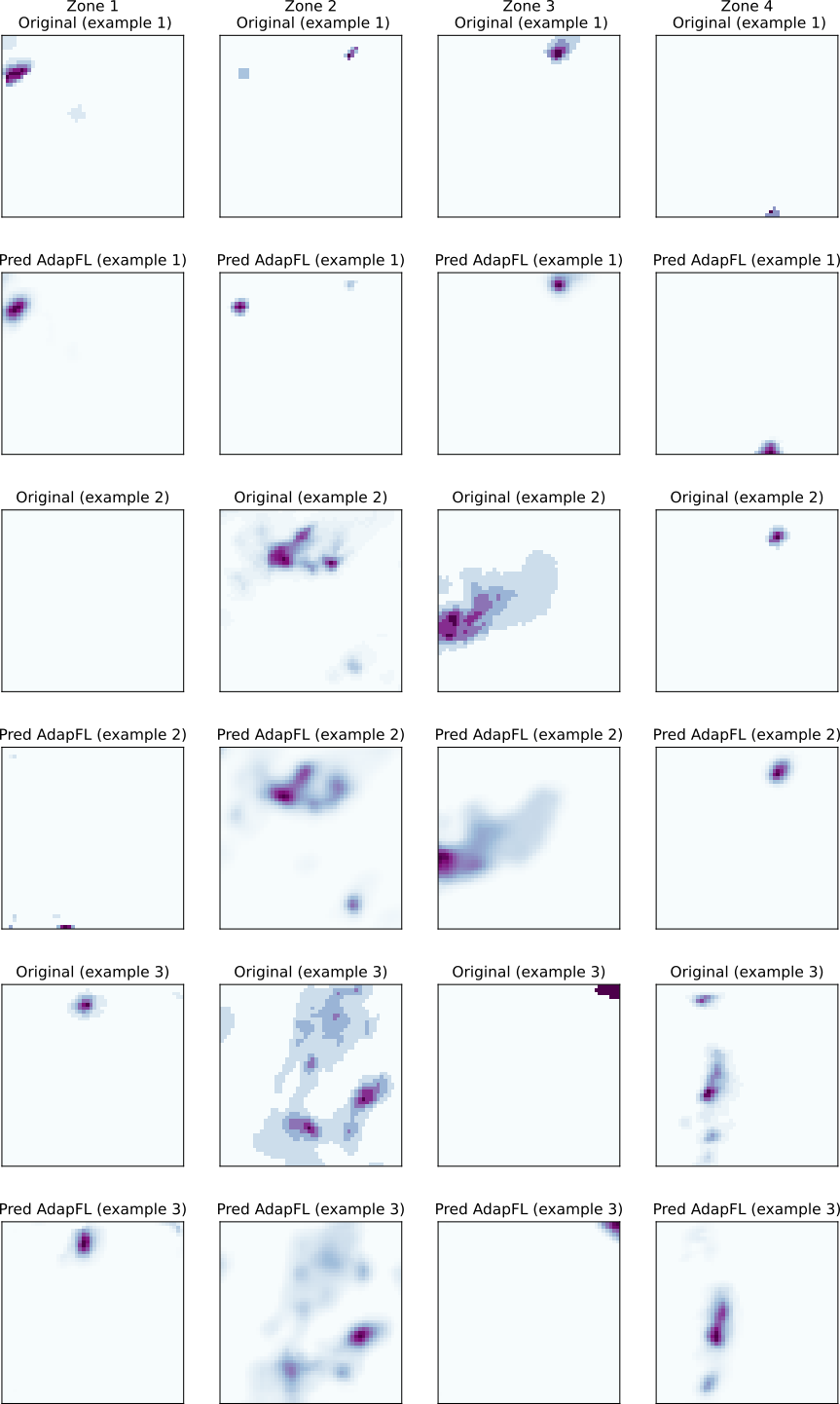}
    \caption{Example of predictions obtained with the adaptive federated learning approach in each of the four zones (1/2).}
    \label{fig:adapfl_pred}
\end{figure}

\begin{figure}[ht!]
    \centering
    \includegraphics[width=0.68\linewidth]{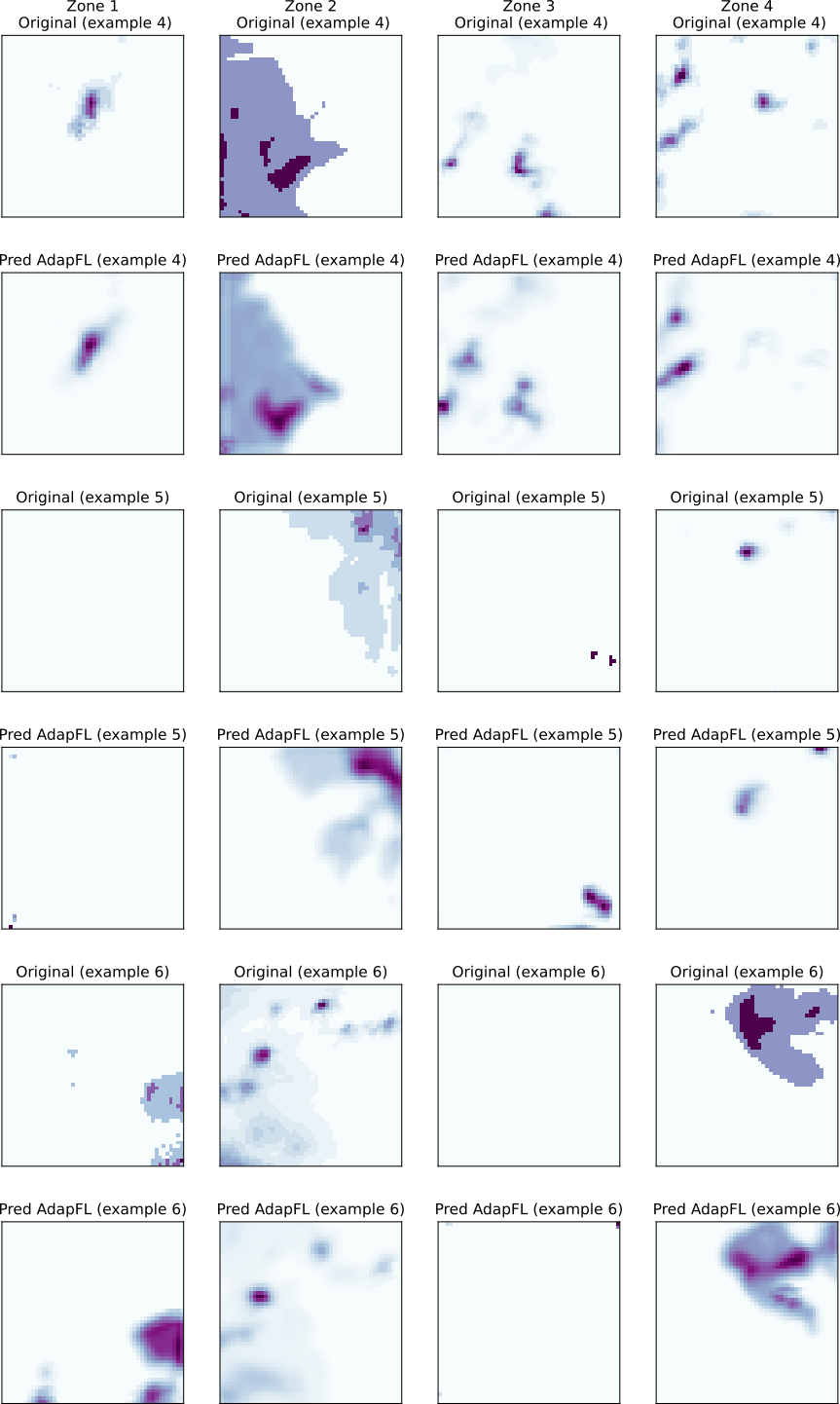}
    \caption{Example of predictions obtained with the adaptive federated learning approach in each of the four zones (2/2).}
    \label{fig:adapfl_pred2}
\end{figure}

Figures~\ref{fig:adapfl_pred} and \ref{fig:adapfl_pred2} show, for each of the four initial artificially distributed zones, six images together with the corresponding predictions obtained with the \textit{adapFL} approach implemented. 
\clearpage

\clearpage
\section{MAE and MSE obtained in each image from the test set}
\label{sec:app_mse_mae_images}

In order to check the robustness of the proposed PFL method, \textit{adapFL}, in this section we show an histogram for the MSE and MAE of each area for each image. We can note that in the vast majority of the images both the MSE and the MAE are near to zero.

\begin{figure}[ht!]
    \centering
    \includegraphics[width=0.9\linewidth]{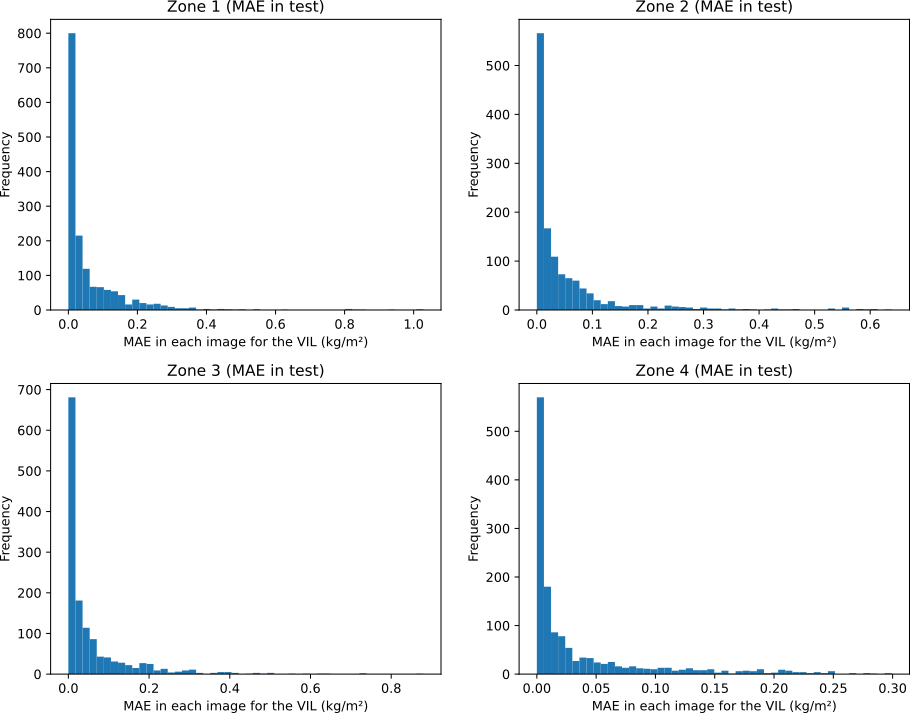}
    \caption{Histogram with the MAE obtained for the images of the test set using the \textit{adapFL} method.}
    \label{fig:adapfl_mae}
\end{figure}
\clearpage
\begin{figure}[ht!]
    \centering
    \includegraphics[width=0.9\linewidth]{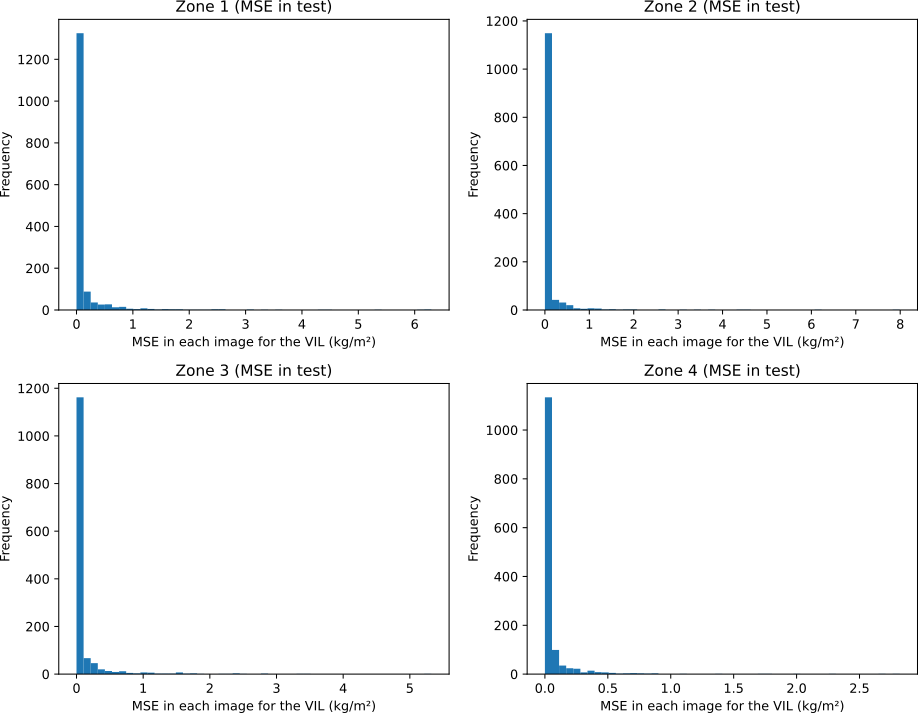}
    \caption{Histogram with the MSE obtained for the images of the test set using the \textit{adapFL} method.}
    \label{fig:adapfl_mse}
\end{figure}

\clearpage
\section{Diagram of the neural network implemented}
\label{sec:app_cnn}
\begin{figure*}[ht]
    \centering
    \includegraphics[width = 0.14\textwidth]{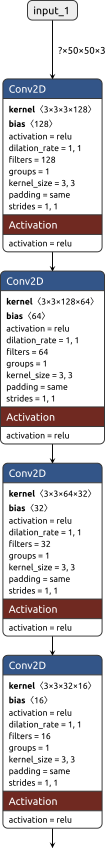}
    \caption{Convolutional neural network implemented (1/2).}
    \label{fig:ANN_uc1}
\end{figure*}

\begin{figure*}[ht]
    \centering
    \includegraphics[width = 0.14\textwidth]{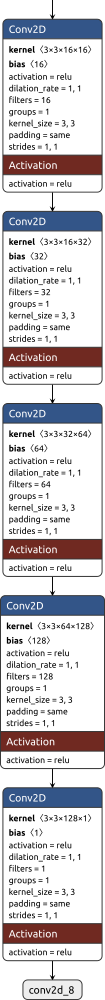}
    \caption{Convolutional neural network implemented (2/2).}
    \label{fig:ANN_uc1_2}
\end{figure*}

The schema of the neural network implemented in this study is shown in Figures~\ref{fig:ANN_uc1} and \ref{fig:ANN_uc1_2}.
\clearpage



\end{appendices}

\newpage
\clearpage
\bibliography{main_arxiv}

\begin{thebibliography}{48}
\providecommand{\natexlab}[1]{#1}
\providecommand{\url}[1]{{#1}}
\providecommand{\urlprefix}{URL }
\providecommand{\doi}[1]{\url{https://doi.org/#1}}
\providecommand{\eprint}[2][]{\url{#2}}
 \bibcommenthead

\bibitem[{Abadi et~al(2015)Abadi, Agarwal, Barham, Brevdo, Chen, Citro, Corrado, Davis, Dean, Devin, Ghemawat, Goodfellow, Harp, Irving, Isard, Jia, Jozefowicz, Kaiser, Kudlur, Levenberg, Man\'{e}, Monga, Moore, Murray, Olah, Schuster, Shlens, Steiner, Sutskever, Talwar, Tucker, Vanhoucke, Vasudevan, Vi\'{e}gas, Vinyals, Warden, Wattenberg, Wicke, Yu, and Zheng}]{tensorflow2015-whitepaper}
Abadi M, Agarwal A, Barham P, et~al (2015) {TensorFlow}: Large-scale machine learning on heterogeneous systems. \urlprefix\url{https://www.tensorflow.org/}, software available from tensorflow.org

\bibitem[{Agrawal et~al(2019)Agrawal, Barrington, Bromberg, Burge, Gazen, and Hickey}]{agrawal2019machine}
Agrawal S, Barrington L, Bromberg C, et~al (2019) Machine learning for precipitation nowcasting from radar images. \eprint{1912.12132}

\bibitem[{Agrawal et~al(2022)Agrawal, Sarkar, Aouedi, Yenduri, Piamrat, Alazab, Bhattacharya, Maddikunta, and Gadekallu}]{AGRAWAL2022346}
Agrawal S, Sarkar S, Aouedi O, et~al (2022) Federated learning for intrusion detection system: Concepts, challenges and future directions. Computer Communications 195:346--361. \doi{https://doi.org/10.1016/j.comcom.2022.09.012}, \urlprefix\url{https://www.sciencedirect.com/science/article/pii/S0140366422003516}

\bibitem[{{AI4EOSC project}(2024)}]{ai4eosc}
{AI4EOSC project} (2024) {AI4EOSC website.} \urlprefix\url{https://ai4eosc.eu/}, [Accessed 19-02-2024]

\bibitem[{Chen et~al(2020)Chen, Cao, Ma, and Zhang}]{chen2020deep}
Chen L, Cao Y, Ma L, et~al (2020) A deep learning-based methodology for precipitation nowcasting with radar. Earth and Space Science 7(2):e2019EA000812

\bibitem[{Cuomo and Chandrasekar(2021)}]{cuomo2021use}
Cuomo J, Chandrasekar V (2021) Use of deep learning for weather radar nowcasting. Journal of Atmospheric and Oceanic Technology 38(9):1641--1656

\bibitem[{Fan and Ling(2018)}]{FanLing2018}
Fan H, Ling H (2018) Siamese cascaded region proposal networks for real-time visual tracking. \eprint{1812.06148}

\bibitem[{Goodfellow et~al(2014)Goodfellow, Pouget-Abadie, Mirza, Xu, Warde-Farley, Ozair, Courville, and Bengio}]{Goodfellow2014}
Goodfellow I, Pouget-Abadie J, Mirza M, et~al (2014) Generative adversarial nets. In: Ghahramani Z, Welling M, Cortes C, et~al (eds) Advances in Neural Information Processing Systems, vol~27. Curran Associates, Inc., \urlprefix\url{https://proceedings.neurips.cc/paper_files/paper/2014/file/5ca3e9b122f61f8f06494c97b1afccf3-Paper.pdf}

\bibitem[{Gu et~al(2018)Gu, Wang, Kuen, Ma, Shahroudy, Shuai, Liu, Wang, Wang, Cai, and Chen}]{GuWang2018}
Gu J, Wang Z, Kuen J, et~al (2018) Recent advances in convolutional neural networks. Pattern Recognition 77:354--377. \doi{10.1016/j.patcog.2017.10.013}

\bibitem[{Han et~al(2023)Han, Shin, Im, and Lee}]{han2023key}
Han D, Shin Y, Im J, et~al (2023) Key factors for quantitative precipitation nowcasting using ground weather radar data based on deep learning. Geoscientific Model Development Discussions 2023:1--43

\bibitem[{Han et~al(2022)Han, Zhao, Chen, and Chandrasekar}]{9354430}
Han L, Zhao Y, Chen H, et~al (2022) Advancing radar nowcasting through deep transfer learning. IEEE Transactions on Geoscience and Remote Sensing 60:1--9. \doi{10.1109/TGRS.2021.3056470}

\bibitem[{Hanel and Buishand(2010)}]{Hanel2010}
Hanel M, Buishand TA (2010) On the value of hourly precipitation extremes in regional climate model simulations. Journal of Hydrology 393(3):265--273. \doi{10.1016/j.jhydrol.2010.08.024}

\bibitem[{Hosseinzadehtalaei et~al(2020)Hosseinzadehtalaei, Tabari, and Willems}]{Hossein2020}
Hosseinzadehtalaei P, Tabari H, Willems P (2020) Climate change impact on short-duration extreme precipitation and intensity–duration–frequency curves over europe. Journal of Hydrology 590:125249. \doi{10.1016/j.jhydrol.2020.125249}

\bibitem[{Jolliffe and Stephenson(2012)}]{jolliffe2012forecast}
Jolliffe IT, Stephenson DB (2012) Forecast verification: a practitioner's guide in atmospheric science. John Wiley \& Sons

\bibitem[{Kesa et~al(2022)Kesa, Styles, and Sanchez}]{Kesa2022}
Kesa O, Styles O, Sanchez V (2022) Multiple object tracking and forecasting: Jointly predicting current and future object locations. In: 2022 IEEE/CVF Winter Conference on Applications of Computer Vision Workshops (WACVW), pp 560--569, \doi{10.1109/WACVW54805.2022.00062}

\bibitem[{Kim et~al(2005)Kim, Fisher, Yezzi, Çetin, and Willsky}]{Kim2005}
Kim J, Fisher JW, Yezzi A, et~al (2005) A nonparametric statistical method for image segmentation using information theory and curve evolution. IEEE Transactions on Image Processing 14(10):1486--1502. \doi{10.1109/TIP.2005.854442}

\bibitem[{Kingma and Ba(2014)}]{kingma2014adam}
Kingma DP, Ba J (2014) Adam: A method for stochastic optimization. arXiv preprint arXiv:14126980

\bibitem[{Ko et~al(2022)Ko, Lee, Hwang, Oh, Son, and Shin}]{KO2022105072}
Ko J, Lee K, Hwang H, et~al (2022) Effective training strategies for deep-learning-based precipitation nowcasting and estimation. Computers \& Geosciences 161:105072. \doi{https://doi.org/10.1016/j.cageo.2022.105072}, \urlprefix\url{https://www.sciencedirect.com/science/article/pii/S009830042200036X}

\bibitem[{Komj{\'a}ti et~al(2022)Komj{\'a}ti, Varga, M{\'e}ri, Breuer, and Kun}]{Komjati2022}
Komj{\'a}ti K, Varga {\'A}J, M{\'e}ri L, et~al (2022) Investigation of a supercell merger leading to the ef4 tornado in the czech republic on june 24, 2021 using radar data and numerical model outputs. Id{\"o}j{\'a}r{\'a}s 126(4)

\bibitem[{Korosec(2021)}]{Korosec2021}
Korosec M (2021) The most powerful tornado on record hit the czech republic, leaving several fatalities and 200+ injured across the hodonin district. Weather Report \urlprefix\url{https://www.severe-weather.eu/weather-report/europe-severe-weather-tornado-hodonin-czech-republic-mk/}, published: 25/06/2021

\bibitem[{Kumar et~al(2024)Kumar, Haral, Kalapureddy, Singh, Yadav, Chattopadhyay, Pattanaik, Rao, and Mohapatra}]{KUMAR2024103600}
Kumar B, Haral H, Kalapureddy M, et~al (2024) Utilizing deep learning for near real-time rainfall forecasting based on radar data. Physics and Chemistry of the Earth, Parts A/B/C 135:103600. \doi{https://doi.org/10.1016/j.pce.2024.103600}, \urlprefix\url{https://www.sciencedirect.com/science/article/pii/S1474706524000585}

\bibitem[{Li et~al(2018)Li, Wu, Wang, Zhang, Xing, and Yan}]{LiWu2018}
Li B, Wu W, Wang Q, et~al (2018) Siamrpn++: Evolution of siamese visual tracking with very deep networks. \eprint{1812.11703}

\bibitem[{McMahan et~al(2017)McMahan, Moore, Ramage, Hampson, and Arcas}]{pmlr-v54-mcmahan17a}
McMahan B, Moore E, Ramage D, et~al (2017) {Communication-Efficient Learning of Deep Networks from Decentralized Data}. In: Singh A, Zhu J (eds) Proceedings of the 20th International Conference on Artificial Intelligence and Statistics, Proceedings of Machine Learning Research, vol~54. PMLR, pp 1273--1282, \urlprefix\url{https://proceedings.mlr.press/v54/mcmahan17a.html}

\bibitem[{Niknam et~al(2020)Niknam, Dhillon, and Reed}]{9141214}
Niknam S, Dhillon HS, Reed JH (2020) Federated learning for wireless communications: Motivation, opportunities, and challenges. IEEE Communications Magazine 58(6):46--51. \doi{10.1109/MCOM.001.1900461}

\bibitem[{Novak(2007)}]{NOVAK2007450}
Novak P (2007) The czech hydrometeorological institute's severe storm nowcasting system. Atmospheric Research 83(2):450--457. \doi{https://doi.org/10.1016/j.atmosres.2005.09.014}, european Conference on Severe Storms 2004

\bibitem[{Pavlík et~al(2022)Pavlík, Rozinajová, and Ezzeddine}]{Pavlik2022}
Pavlík P, Rozinajová V, Ezzeddine AB (2022) Radar-based volumetric precipitation nowcasting: A 3d convolutional neural network with u-net architecture. In: 2nd Workshop on Complex Data Challenges in Earth Observation (CDCEO 2022), \urlprefix\url{https://ceur-ws.org/Vol-3207/paper10.pdf}

\bibitem[{Pfitzner et~al(2021)Pfitzner, Steckhan, and Arnrich}]{10.1145/3412357}
Pfitzner B, Steckhan N, Arnrich B (2021) Federated learning in a medical context: A systematic literature review. ACM Trans Internet Technol 21(2). \doi{10.1145/3412357}, \urlprefix\url{https://doi.org/10.1145/3412357}

\bibitem[{Rajczak and Schär(2017)}]{Rajczak2017}
Rajczak J, Schär C (2017) Projections of future precipitation extremes over europe: A multimodel assessment of climate simulations. Journal of Geophysical Research: Atmospheres 122(20):10,773--10,800. \doi{10.1002/2017JD027176}

\bibitem[{Ravuri et~al(2021)Ravuri, Lenc, Willson, Kangin, Lam, Mirowski, Fitzsimons, Athanassiadou, Kashem, Madge, Prudden, Mandhane, Clark, Brock, Simonyan, Hadsell, Robinson, Clancy, Arribas, and Mohamed}]{Ravuri2021}
Ravuri S, Lenc K, Willson M, et~al (2021) Skilful precipitation nowcasting using deep generative models of radar. Nature 597:672–677. \doi{https://doi.org/10.1038/s41586-021-03854-z}

\bibitem[{Rieke et~al(2020)Rieke, Hancox, Li, Milletari, Roth, Albarqouni, Bakas, Galtier, Landman, Maier-Hein et~al}]{rieke2020future}
Rieke N, Hancox J, Li W, et~al (2020) The future of digital health with federated learning. NPJ digital medicine 3(1):119

\bibitem[{Rinehart and Garvey(1978)}]{Rinehart1978}
Rinehart R, Garvey E (1978) Three-dimensional storm motion detection by conventional weather radar. Nature 273(5660):287--289

\bibitem[{Sabah et~al(2024)Sabah, Chen, Yang, Azam, Ahmad, and Sarwar}]{SABAH2024122874}
Sabah F, Chen Y, Yang Z, et~al (2024) Model optimization techniques in personalized federated learning: A survey. Expert Systems with Applications 243:122874. \doi{https://doi.org/10.1016/j.eswa.2023.122874}, \urlprefix\url{https://www.sciencedirect.com/science/article/pii/S0957417423033766}

\bibitem[{S{\'a}inz-Pardo~D{\'\i}az and L{\'o}pez~Garc{\'\i}a(2023)}]{sainzpardo2023fl}
S{\'a}inz-Pardo~D{\'\i}az J, L{\'o}pez~Garc{\'\i}a {\'A} (2023) Study of the performance and scalability of federated learning for medical imaging with intermittent clients. Neurocomputing 518:142--154

\bibitem[{Shi et~al(2015)Shi, Chen, Wang, Yeung, Wong, and Woo}]{shi2015convolutional}
Shi X, Chen Z, Wang H, et~al (2015) Convolutional lstm network: A machine learning approach for precipitation nowcasting. Advances in neural information processing systems 28

\bibitem[{Shi et~al(2017)Shi, Gao, Lausen, Wang, Yeung, Wong, and WOO}]{NIPS2017_a6db4ed0}
Shi X, Gao Z, Lausen L, et~al (2017) Deep learning for precipitation nowcasting: A benchmark and a new model. In: Guyon I, Luxburg UV, Bengio S, et~al (eds) Advances in Neural Information Processing Systems, vol~30. Curran Associates, Inc., \urlprefix\url{https://proceedings.neurips.cc/paper_files/paper/2017/file/a6db4ed04f1621a119799fd3d7545d3d-Paper.pdf}

\bibitem[{Svoboda and Pekárová(1998)}]{Svoboda1998}
Svoboda A, Pekárová P (1998) The catastrophic flood of july 1998 in the malá svinka catchment – simulation of its course [in slovak]. Vodohospodársky časopis 46(6):356–365

\bibitem[{Svoboda et~al(3916)Svoboda, Hanel, Máca, and Kyselý}]{Svoboda2016}
Svoboda V, Hanel M, Máca P, et~al (3916) Projected changes of rainfall event characteristics for the czech republic. Journal of Hydrology and Hydromechanics 64(4):415--425. \doi{doi:10.1515/johh-2016-0036}

\bibitem[{{Sáinz-Pardo Díaz} et~al(2023){Sáinz-Pardo Díaz}, Castrillo, and Álvaro {López García}}]{SAINZPARDODIAZ2023120726}
{Sáinz-Pardo Díaz} J, Castrillo M, Álvaro {López García} (2023) Deep learning based soft-sensor for continuous chlorophyll estimation on decentralized data. Water Research 246:120726. \doi{https://doi.org/10.1016/j.watres.2023.120726}, \urlprefix\url{https://www.sciencedirect.com/science/article/pii/S0043135423011661}

\bibitem[{Tan et~al(2022)Tan, Yu, Cui, and Yang}]{tan2022towards}
Tan AZ, Yu H, Cui L, et~al (2022) Towards personalized federated learning. IEEE Transactions on Neural Networks and Learning Systems

\bibitem[{Tanaka(2002)}]{Tanaka2002}
Tanaka K (2002) Statistical-mechanical approach to image processing. Journal of Physics A: Mathematical and General 35(37):R81. \doi{10.1088/0305-4470/35/37/201}

\bibitem[{Tang and Matyas(2018)}]{Tang2018}
Tang J, Matyas C (2018) A nowcasting model for tropical cyclone precipitation regions based on the trec motion vector retrieval with a semi-lagrangian scheme for doppler weather radar. Atmosphere 9(5). \doi{10.3390/atmos9050200}

\bibitem[{Wang et~al(2019)Wang, Zhang, Bertinetto, Hu, and Torr}]{WangZhang2019}
Wang Q, Zhang L, Bertinetto L, et~al (2019) Fast online object tracking and segmentation: A unifying approach. \eprint{1812.05050}

\bibitem[{Woo and Wong(2017)}]{WooWong2017}
Woo Wc, Wong Wk (2017) Operational application of optical flow techniques to radar-based rainfall nowcasting. Atmosphere 8(3). \doi{10.3390/atmos8030048}

\bibitem[{Yang et~al(2019)Yang, Zhang, Ye, Li, and Xu}]{yang2019ffd}
Yang W, Zhang Y, Ye K, et~al (2019) Ffd: A federated learning based method for credit card fraud detection. In: Big Data--BigData 2019: 8th International Congress, Held as Part of the Services Conference Federation, SCF 2019, San Diego, CA, USA, June 25--30, 2019, Proceedings 8, Springer, pp 18--32

\bibitem[{Zhang et~al(1988)Zhang, Tanida, Itoh, and Ichioka}]{Zhang1988}
Zhang W, Tanida J, Itoh K, et~al (1988) Shift-invariant pattern recognition neural network and its optical architecture. In: Proceedings of annual conference of the Japan Society of Applied Physics, Montreal, CA

\bibitem[{Zhang et~al(2021)Zhang, Wang, Zhou, Wu, and Zhang}]{fl_noniid}
Zhang W, Wang X, Zhou P, et~al (2021) Client selection for federated learning with non-iid data in mobile edge computing. IEEE Access 9:24462--24474. \doi{10.1109/ACCESS.2021.3056919}

\bibitem[{Zhang and Peng(2019)}]{ZhangPeng2019}
Zhang Z, Peng H (2019) Deeper and wider siamese networks for real-time visual tracking. \eprint{1901.01660}

\bibitem[{Zhu et~al(2018)Zhu, Wang, Li, Wu, Yan, and Hu}]{ZhuWang2018}
Zhu Z, Wang Q, Li B, et~al (2018) Distractor-aware siamese networks for visual object tracking. \eprint{1808.06048}

\end{thebibliography}

\end{document}